\documentclass[final,5pt,times,twocolumn]{elsarticle}
\usepackage{llm4kg-2023}

\journal{Knowledge-Based Systems}

\begin{document}

\begin{frontmatter}

\title{Large language models as oracles for instantiating ontologies\\with domain-specific knowledge}

\author[disi]{Giovanni Ciatto}
\ead{giovanni.ciatto@unibo.it}

\author[disi]{Andrea Agiollo}
\ead{andrea.agiollo@unibo.it}

\author[disi]{Matteo Magnini}
\ead{matteo.magnini@unibo.it}

\author[disi]{Andrea Omicini}
\ead{andrea.omicini@unibo.it}


\affiliation[disi]{
    organization={Department of Computer Science and Engineering (DISI), Alma Mater Studiorum -- Università di Bologna},
    addressline={via dell'Università 50},
    city={Cesena, FC},
    postcode={47522},
    country={Italy}
}

\begin{abstract}
\inlineParagraph{Background}
Endowing intelligent systems with semantic data commonly requires
designing and instantiating ontologies with domain-specific knowledge.
Especially in the early phases,
those activities are typically performed \emph{manually} by human experts possibly leveraging on their own experience.
The resulting process is therefore time-consuming, error-prone, and often biased by the personal background of the ontology designer.
\inlineParagraph{Objective}
To mitigate that issue,
we propose a novel domain-independent approach to automatically instantiate
ontologies with domain-specific knowledge,
by leveraging on large language models (LLMs) as oracles.
\inlineParagraph{Method}
Starting from
\begin{inlinelist}
    \item an initial schema composed by inter-related classes and properties
    and
    \item a set of query templates,
\end{inlinelist}
our method queries the LLM multiple times,
and generates instances for both classes and properties from its replies.
Thus,
the ontology is \emph{automatically} filled with domain-specific knowledge,
compliant to the initial schema.
As a result,
the ontology is quickly and automatically enriched with manifold instances,
which experts may consider to keep, adjust, discard, or complement
according to their own needs and expertise.
\inlineParagraph{Contribution}
We formalise our method in general way and instantiate it over various LLMs,
as well as on a concrete case study.
We report experiments rooted in the nutritional domain
where an ontology of food meals and their ingredients is automatically instantiated from scratch,
starting from a categorisation of meals and their relationships.
There,
we analyse the quality of the generated ontologies and compare ontologies attained by exploiting different LLMs.
Experimentally, our approach achieves a quality metric that is up to five times higher than the state-of-the-art, while reducing erroneous entities and relations by up to ten times.
Finally, we provide a SWOT analysis of the proposed method.
%

\end{abstract}



\begin{keyword}
ontology population \sep large language models \sep nutrition \sep automation \sep domain-specific knowledge
\end{keyword}

\end{frontmatter}

\section{Introduction}

Nowadays,
the demand for intelligent systems
capable of understanding, reasoning, and interacting with complex information
is becoming paramount.
However,
the possibility of truly intelligent systems
hinges upon the incorporation of \emph{machine-interpretable} knowledge representations
that can capture humans' \emph{domain-specific} knowledge from diverse domains.

The rapid proliferation of data and data representation formats,
coupled with the need for intelligent systems to comprehend and contextualise ambiguous, vague, and often incomplete information,
brings about the question
``how can we precisely represent domain-specific information in a machine-understandable manner?'',
whose answer is rooted in the notion of \emph{ontology}~\cite{Grimm10}.

Regardless of the particular technological reification of choice,
ontologies are formal,
extensional representations of knowledge following the principles of description logics.
There,
concepts and relationships among concepts are formally defined and expressed in a machine-readable format,
facilitating computational systems to reason about the world in a manner akin to human cognition.
In other words,
ontologies provide a means to convey human knowledge in a machine-understandable way.

Unlike sub-symbolic approaches that rely on distributed or loosely organised data~\cite{skeislr-acmcs},
ontologies provide a structured framework that defines concepts, relationships, and properties within a specific domain.
Such a structured knowledge representation is a key enabling factor in bridging the semantic gap,
pushing AI systems beyond superficial pattern recognition
and possibly grasp the underlying meaning of the data.

Hence,
from the human perspective,
ontologies provide a shared vocabulary and an unambiguous understanding of domain-specific knowledge.
This comes at the cost of a meticulous and time-consuming process of ontology creation,
which should be performed in such a way to guarantee
some degree of adherence among the ontology
and the real-world domain it represents.

%

Currently,
ontologies are populated either manually,
by domain experts or communities,
or through semi-automatic extraction from data~\cite{petasis-2011}.
On the one hand,
manual ontology population is time- and effort-consuming,
and possibly affected by humans' errors and biases.
However,
if the population process is long and inclusive enough,
it may easily yield higher-quality
(in terms of adherence to reality)
results on the long run.
On the other hand,
data-driven approaches offer speed and scalability
but might yield lower-quality ontologies.
In this case,
low quality may be due to biases in the data or to the characteristic of the extraction procedure
(e.g.\ data may be under-representing some aspects of the domain,
or the procedure may trade off completeness for speed).

Arguably,
the ideal population procedure would involve both human- and data-driven approaches:
the latter should provide ``mass'' to the ontology,
while the former should refine and revise its details.
Along this line,
this paper introduces \llmfkg:
a novel approach that aims at amalgamating the strengths of both human- and data-driven ontology population methods.

Leveraging the observation that
large language models (LLM) are trained on various data from the entire web,
we hypothesise that these models encapsulate a substantial amount of domain-specific knowledge.
Based on this premise,
we propose an automated procedure,
\llmfkg,
to extract domain-specific knowledge from LLM
and use it for populating ontologies.
Starting from
\begin{inlinelist}
    \item an initial schema composed by inter-related classes and properties,
    and
    \item a set of query templates
\end{inlinelist}
our method queries the LLM multiple times,
and it generates instances for both classes, relationships, and properties from its replies.
Further queries are performed to refine the ontology,
and to balance the instances in the class hierarchy.
In this way,
the ontology is \emph{automatically} filled with domain-specific knowledge from the LLM,
while remaining compliant to the initial schema.
As result,
the ontology is quickly and automatically enriched with manifold instances,
which experts may consider to keep, adjust, discard, or complement
according to their own needs and expertise.

While data-driven,
our method comes with several peculiarities w.r.t.\ other state-of-the-art approaches.
First,
it is not tailored upon any specific dataset,
but rather it uses LLM as oracles to generate data.
%
%
Second,
it supports not only the population of the ontology,
but its refinement as well.
In fact,
it is incremental,
in the sense that it can be applied to already-instantiated ontologies,
hence enriching or populating them even more.
Finally,
it is general-purpose,
in the sense that it can be applied to virtually any domain,
and to different LLM.

To validate our approach,
we provide a Python implementation of \llmfkg,
and we let it populate a custom OWL ontology of ours
and we evaluate the quality of the resulting ontology.
The evaluation involves a thorough inspection of the populated instances,
where we manually and carefully assess whether
LLM-generated individuals are meaningful
and correctly placed within the ontology structure.
We repeat the experiment on different LLMs,
and we compare the results.

Broadly speaking,
our method contributes to the ongoing discourse on ontology population methods,
presenting a promising hybrid approach that harnesses the power of LLMs
to enhance the efficiency and accuracy of the ontology creation process.
%
%
Along this line,
to deepen the discussion about the strenghts, weaknesses, opportunities, and threats of our method,
we dedicate an entire section to a SWOT analysis of our approach.

Accordingly,
the remainder of this paper is organised as follows.
\Cref{sec:background} provides background on ontologies, LLMs, and related work.
\Cref{sec:contribution} presents the \llmfkg{} method. 
%
\Cref{sec:case-study} introduces a case study 
where \llmfkg{} is validated, and results from different LLM families are compared.
After \Cref{sec:discussion} reports a SWOT analysis of our method,
\Cref{sec:conclusions} concludes the paper.

\section{Background}
\label{sec:background}

In this section,
we recall basic theoretical and practical notions about ontologies, 
and we summarise the state of the art in ontology population.
Then,
we provide an overview on large language models. 

\subsection{Ontologies, Description Logics, and Semantic Web}

An ontology is ``a formal explicit specification of a shared conceptualisation of a domain
of interest''~\cite{Grimm10}.
It describes the entities involved in a specific domain, and their properties, as well as the relationships among them.
Inside an ontology, information is organised and represented in such a way that it can be interpreted by both humans
and machines.

In this paper,
we adopt the syntax and semantics of
the well-known \alc{} description logics~\cite{BaaderHS08}.
In \alc{}, 
an ontology 
is
a set of axioms aimed at specifying:
\begin{description}
    \item[classes] (a.k.a. \textbf{concepts}) 
    are set of entities of the same sort.
    %
    %
    We denote classes either by their mnemonic names,
    in capital italics
    (e.g.: $\concept{Animal}$) 
    or by expressions composing simpler classes via algebraic operations
    such as union ($\sqcup$), intersection ($\sqcap$), or negation ($\neg$).

    \item[instances] (a.k.a. \textbf{individuals}) 
    are constants representing (identifiers of) relevant entities.
    %
    %
    We denote instances by mnemonic names 
    in lowercase monospaced font, e.g.: $\instance{tom}$.
    Entities may be part of one or more classes:
    we use operator $:$ to denote the ``is a''
    (a.k.a., ``instance of'')
    relation,
    %
    e.g., ``$\instance{tom} : \concept{Cat}$'' $\equiv$ ``Tom is a cat''.

    \item[properties] (a.k.a. \textbf{roles}) 
    are binary relations among pairs of instances.
    %
    %
    We denote properties by mnemonic names 
    in lowercase sans-serif font, e.g.\ $\relation{eats}$.
    As binary relations,
    properties require specifying which pair of concepts they are linking together.
    %
    For example,
    we write $\relation{eats} \subseteq \concept{Animal} \times \concept{Edible}$ meaning that
    one animal may, in general, eat something edible.
    In that case,
    we say that $\concept{Animal}$ is the \emph{domain} of $\relation{eats}$, and $\concept{Edible}$ is
    its \emph{range}.
    For any two actual instances in the relation,
    we use the prefix functional notation to denote that the
    relationship holds among those two instances,
    e.g.\ $\relation{eats}(\instance{tom}, \instance{jerry})$.
\end{description}
Both concepts and roles may be subject to the \textbf{subsumption}
(a.k.a. \textbf{subclass})
relation ($\sqsubseteq$),
which essentially corresponds to set inclusion.
So,
for instance,
$\concept{Cat} \sqsubseteq \concept{Animal}$ states that all cats are also animals,
whereas
$\relation{predatorOf} \sqsubseteq \relation{eats}$ means that,
for any couple of animals, if one eats another, then the former is a predator for the latter.

Two special concepts are introduced in the \alc{} logic,
namely $\top$ and $\bot$\,---%
respectively, the set of all instances and the set with no instances.
These are used to denote the most general and the most specific concepts,
respectively.
%
%
Thanks to subsumption,
concepts (and properties) can be organised hierarchically,
forming a taxonomy%
---i.e. a \emph{graph} where nodes are concepts and arcs are subsumption relations.

Summarising, any \alc{} theory would consist of
\begin{inlinelist}
    \item \emph{terminological} axioms (a.k.a. TBox),
    aimed at defining concepts and properties,
    as well subsumption or equivalence relations among them,
    and
    \item \emph{assertion} axioms (a.k.a. ABox),
    aimed at assigning instances to either classes or properties.
\end{inlinelist}
When it comes to practice, and a new ontology is being defined from scratch,
the TBox is usually defined first,
in the \emph{modelling} phase,
and the ABox is defined next,
in the \emph{population} phase.
Both phases could be iterated multiple times
until the ontology is deemed complete and satisfactory,
but the key point is:
there is a modelling phase,
where concepts and properties are defined,
and a population phase,
where instances are assigned to them.

Technologically speaking,
ontologies are most often expressed by means of Semantic Web~\cite{bernerslee2001semantic} languages,
such as the Web Ontology Language\footnote{\url{https://www.w3.org/OWL/}} (OWL)
or the Resource Description Framework\footnote{\url{https://www.w3.org/2001/sw/wiki/RDF}} (RDF).
These languages come along with concrete syntaxes for representing ontologies,
as well as with software tools facilitating their manipulation, visualisation, and processing.
Among these,
we rely upon Protégé~\cite{Musen15} for the manual editing and inspection of ontologies,
and Owlready2\footnote{\url{https://github.com/pwin/owlready2}}~\cite{Lamy17} for automating
ontology manipulation by means of the Python language.

\subsection{Related works on Ontology Population}

The problem of filling ontologies with instances is well-known in the Semantic Web community~\cite{lubani-2019}.
%
It
consists of expanding an ontology with further assertions
(i.e., ABox axioms),
compliant with the concepts and properties therein defined
(i.e., TBox axioms).
Such problem is commonly referred to in the literature as ``ontology population''%
---or ``learning'',
when the authors want to stress that the ontology is being populated via some data-driven (semi-)automatic procedure~\cite{petasis-2011}.
Some authors have also used the expression ``ontology learning'' to refer to
the process of (semi-)automatically inferring the whole ontology
-- there including TBox axioms too --,
hence why we stick to the ``ontology population'' nomenclature.

The need of a (semi-)automatic procedure to populate ontologies is motivated by the fact that
\emph{manual} population activities are time-consuming and error-prone,
and therefore inherently costly~\cite{cherifa-2021}.

Several methods for structured data extraction from text have been proposed in the literature
to serve the purpose of semi-automatic ontology population.
Broadly speaking,
these methods can be classified into two main categories:
\emph{linguistics-based}
and
\emph{machine-learning} (ML) approaches~\cite{Alaswadi-2020}.

Linguistics-based approaches~\cite{finkestein-1999,morin-1999,harith-2003} are early techniques,
including, but not limited to:
analysis of the syntactic structure of sentences,
pattern-based extraction,
part-of-speech tagging,
dictionaries,
etc.
Because these approaches are not fully automatic,
they still need lots of human effort 
(e.g., corpora selection, supervision, revision, parameter fine-tuning, etc.).
Moreover,
these methods are often tightly coupled the domain they have been designed for,
(e.g., different corpora might lead to different statistical information for the same terms,
therefore different thresholds may be required).

ML approaches may either rely on statistic-based shallow ML,
or deep learning.
Statistic-based approaches attempt to convert free text into numerical data
by relying on classical NLP techniques
(e.g. term frequency-inverse document frequency, TF-IDF)
which is
later used to feed shallow data-mining algorithms
(e.g. decision trees).
%
The main limitation of these methods 
is that they may struggle to capture contextual information from the text.
The interested reader may find examples of methods from this category in the following
works~\cite{tanev-2006,yoon-2007,maynard-2008,celjuska-2004,etzioni-2005,jiang-2011,souili-2015}.

Recent trends in ontology population involve the application of deep learning techniques~\cite{liu-2013,zeng-2014,ayadi-2019}.
Differently than shallow ML approaches,
which rely on hand-crafted, statistical representations of text,
deep learning approaches attempt to learn such representations automatically,
from the data. 
%
In this way,
contextual information in the text is more likely to be captured.

Overall,
all the approaches mentioned so far require a large corpus of textual documents to be available,
from which concepts, instances, and properties are extracted.
This is a major limitation,
because it requires the ontology designer to either manually collect the documents,
or to rely on third-party document sources.
In both cases,
the corpus of documents may be biased,
incomplete,
non-representative of the domain,
or simply lacking.

The reliance on large amounts of data also implies that
state-of-the-art population methods are very domain-sensitive,
and poorly incremental.
Here,
domain-sensitivity refers to the availability of documents for the domain of interest:
if the domain is too narrow,
it may be difficult to find enough relevant documents to populate the ontology;
if the domain is too broad,
it may be difficult to avoid the inclusion of irrelevant information in the ontology.
Poor incrementality,
instead,
refers to the fact that,
once the ontology is populated from a given corpus of documents,
it may be difficult to add further instances to the ontology,
without having to re-populate it from scratch.

\subsection{Large Language Models}
\label{sec:llm}

Large-language models are deep NN with enormous number of weights
-- from BERT~\cite{bert-2019} with 340 millions weights
to GPT-3~\cite{gpt3-2020} with 175 billions weights --,
which are specialised on NLP tasks,
and,
in particular,
text generation.
%
%
In this subsection,
we provide a brief overview of LLM,
referencing relevant literature on the topic
and outlining the most relevant aspects that we exploit in our work.
%

\paragraph{The role of pretraining}

One key aspect of modern LLM is that they are \emph{pre-trained} on very large corpora of textual data,
to learn \emph{general-purpose} language models~\cite{MinRSSVNAHR23}.
Language models are sub-symbolic probabilistic representations of natural language,
capturing underlying meaning of words in texts.
%

Before LLM were introduced,
language models were trained on small domain-specific datasets,
resulting in neural networks which were only able to operate in that domain.
Changing the domain implied re-training the model from scratch,
on a possibly different dataset.
This is not the case for LLM,
which are trained 
to learn a reusable language model,
which is in principle-domain agnostic. %

From an engineering perspective,
pre-training is highly beneficial,
as it implies a paradigm shift~\cite{MinRSSVNAHR23}.
%
The data-provisioning, pre-processing, and training pipeline is factorised,
and performed once and for all domains,
while fine-tuning upon need.


\paragraph{Learning from the Web}


LLMs are essentially very big knowledge bases~\cite{PetroniRRLBWM19},
storing the text they have been trained upon,
and supporting efficient information retrieval,
via free-text.
The training text most commonly consists of a fairly wide portion of the publicly available Web.

For instance,
the early BERT model~\cite[sec.\ 3.1]{bert-2019} was trained on the well-known BooksCorpus dataset
(cf.\ \cite[sec.\ 3]{ZhuKZSUTF15})
and on dump of the English Wikipedia of the time.

Similarly,
models from the GPT family~\cite{gpt2-2019, gpt3-2020} were trained on documents scraped from the public Internet,
using Reddit as the entry point~\cite[sec.\ 2.1]{training-data-attack-2021}.

Models following the RoBERTa~\cite{roberta-2019} approach were trained using
\begin{inlinelist}
    \item the same datasets as BERT,
    \item a sample of the Common~Crawl\footnote{\url{https://commoncrawl.org}} News dataset, 
    %
    \item the OpenWebTextCorpus~\cite{Gokaslan2019OpenWeb}, and
    %
    \item the Stories dataset~\cite{TrieuQuoc2018}. 
\end{inlinelist}

The LLAMA model family~\cite{llama1-2023,llama2-2023} relies on a composition of datasets, including
the Colossal Clean Crawled Corpus (C4) \cite{RaffelSRLNMZLL20} 
English Wikipedia,
and many more.
%

Finally,
according to their Web site\footnotemark,
MistralAI's Mixtral-of-Experts model is ``pre-trained on data extracted from the open Web''.
\footnotetext{\url{https://mistral.ai/news/mixtral-of-experts}}

\paragraph{Oracles for data generation}

%
Our hypothesis is that Web-wise pre-training lets LLM accumulate substantial amounts of domain-specific knowledge,
on manifold domains%
---virtually, all the ones described on the Web.
Such knowledge may then be extracted by querying the LLM with adequate prompts,
aimed at letting the LLM generate relevant text for the domain at hand.
In the perspective of the user,
this implies that LLM may be exploited as \emph{oracles} for information retrieval
tasks.


\paragraph{Hallucinations}

It is worth mentioning that,
at the current state of technology,
LLMs are not perfect oracles,
and they may generate text that is not coherent with the context of the conversation,
or simply factually wrong.
Such phenomenon is known as \emph{hallucination}~\cite{hallucination-2023},
and it is a major issue to take into account when using LLM as oracles
---as it implies that the LLM cannot be trusted blindly.

\paragraph{Temperature}
\label{par:temperature}

Arguably,
hallucinations 
may be critical if information retrieval is performed by inexperienced users, 
but they may be tolerable if the goal is data generation%
---as some degree of error may be acceptable.
In any case,
some degree of control may be desirable to tolerate the issue.

To mitigate the hallucination issue,
LLM technologies
come with a parameter
called \emph{temperature},
which regulates the randomness of the response.
In the intention of designers,
this should allow users to control the ``creativity'' of the LLM,
and hence to mitigate hallucination.
For instance,
in GPT,
the temperature parameter is a real value between 0 and 1,
regulating the percentage of randomness in the response%
\footnote{
    How does temperature work?
    Let us recall that a language model is essentially a probabilistic representation of which words are likely to
    appear next in the sentence, given words that are currently in the sentence.
    When generating a sentence, the LLM will proceed by sampling the next word from the probability distribution of the
    words that are likely to appear next.
    Setting temperature to 0 implies that the LLM will always choose the most likely word to appear next -- hence
    making text generation completely deterministic --, while setting temperature to 1 implies that the LLM will choose
    the next word randomly, according to the probability distribution.
}

\paragraph{The role of prompts}

Other mitigation techniques may be applied to the hallucination issue,
such as \emph{prompt engineering}.
In fact,
while of course writing clear and unambiguous prompts may help the LLM in matching users' expectations,
it has been observed that the prompt may be tweaked in finer ways.
For instance,
the authors of \cite{MemmertCB24} observed that
explicitly telling the LLM to put itself in the shoes of a domain expert 
may lead to more accurate answers.
%


\paragraph{Emergent capabilities}

In the future, 
we expect LLM performance to improve,
and hallucinations 
to become more and more controllable%
---if not negligible.

Recent research suggests that LLM may be capable of \emph{emergent} capabilities%
---i.e. they learn to do tasks that they have not been trained for.
%
According to Wei et al.~\cite{WeiTBRZBYBZMCHVLDF22},
emergent capabilities are tied to the amount of parameters in the LLM.


Despite some authors hypothesise that
emergent capabilities may just be the result of non-linear evaluation metrics~\cite{SchaefferMK23},
and other authors~\cite{WuQRAKKA23} observed performance degradation in unusual tasks, 
most researchers and practitioners are currently optimistic about future performance improvements of LLM.

\paragraph{Pricing and rating plans}

Using LLM implies having access to considerable computational resources.
Operating LLM
-- even if in inference-only mode --
is currently unfeasible on commodity hardware,
and it requires the use of specialised hardware,
such as GPUs or TPUs. 
%

\begin{figure}
    \centering
    \includegraphics[width=\linewidth]{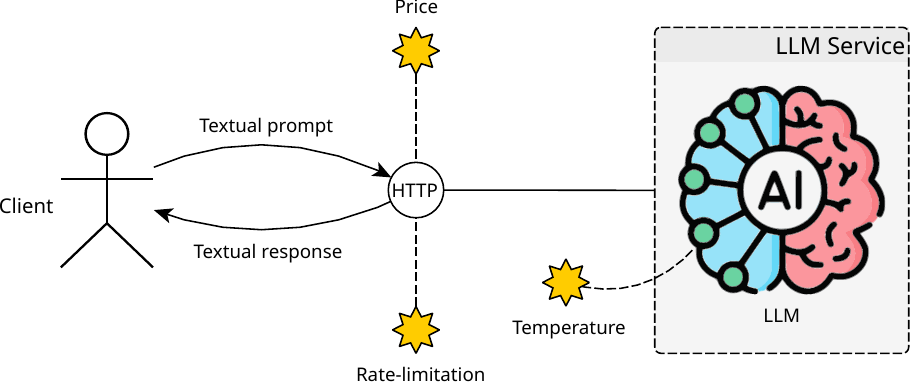}
    \caption{Conceptualization of LLM services}
    \label{fig:llm-concept}
\end{figure}

For this reason,
most LLM technologies nowadays are available ``as-a-Service'', via Web API.
This situation is conceptualised in \Cref{fig:llm-concept}.
%

Two notable examples of LLM technologies available as-a-Service are OpenAI Platform's API\footnotemark
\footnotetext{\url{https://platform.openai.com/}}
-- which provides remote access to GPT-* models --,
and Hugging Face\footnote{\url{https://huggingface.co/}}%
---that is, a Web community where LLM models can be published,
and possibly queried via a unified Web API.
%
which make it easy to automate LLM queries.


One key limitation implied by LLM as Web services open to the general audience is guaranteeing fair access to the service%
---as well as providers covering their operational costs.
For these reasons,
most providers will require users to pay for the service,
or be subject to some rate limitations
when using the service.

For instance,
OpenAI Platform comes with a pricing model\footnote{\url{https://openai.com/pricing}} which depends on
\begin{inlinelist}
    \item the model being queried,
    \item the amount of input tokens in the query,
    and
    \item the amount of output tokens generated by the LLM.
\end{inlinelist}
%
%
That pricing model brings about an implicit incentive on the user side,
towards minimising the amount of tokens in the query,
and in the response, as well.
Similarly,
OpenAI Platform imposes rate constraints\footnotemark{} which depend on
\footnotetext{\url{https://platform.openai.com/docs/guides/rate-limits}}
\begin{inlinelist}
    \item the model being queried,
    \item on the sort of client querying it (paying or free),
    \item on the amount of tokens per request,
    and
    \item requests per time unit from the same client.
\end{inlinelist}
Such rate-limiting model forces the user to artificially slow down any client application querying the LLM in automatic
ways.

Hugging Face comes with similar pricing\footnote{\url{https://huggingface.co/pricing}}
and rate-limiting models\footnote{\url{https://huggingface.co/docs/api-inference/faq\#rate-limits}},
which are however more flexible,
as they include a free -- yet very slow and limited -- option.


\subsection{Related works on LLMs and Knowledge Graphs}
\label{ssec:llm-kg}

In the literature,
there are many works where LLMs are applied to (semi-)automatic knowledge graph (KG) manipulation~\cite{roadmap-kg-2024,ZhuWCQOYDCZ24,PanRKSCDJO0LBMB23},
in ways which are apparently similar to our work.
In this section,
we summarise the most relevant works in this area,
discussing how they differ from ours.

Similarly to ontologies,
KGs represent knowledge in a structured way,
by means of nodes and edges.
Nodes (resp. edges) represent entities (resp. relations).
They are made of triplets (a.k.a. facts) of the form $(s, p, o)$,
where $s$ is the subject,
$p$ is the predicate,
and $o$ is the object.
%
%
However,
differently from ontologies~\cite{kg-vs-ontology-2016},
KGs do not impose any further constraint on the structure of the knowledge they represent.
Conversely,
ontologies use a well-defined set of axioms,
which make them amenable to automated reasoning.
%

In practice,
KGs are more flexible than ontologies,
as they can be populated with triples that do not necessarily respect any pre-defined schema.
Vice versa,
populating an ontology means that,
at the end of the process,
the final ontology is still compliant with the original schema.
%
%
In other words,
KGs can be considered as one possible means
-- among many --
to represent ontologies.

Works from the literature applying LLMs to KGs can be classified into two main categories,
namely:
LLM-augmented KG \emph{completion} and \emph{construction}. 

\paragraph{LLM-augmented KG completion}

KG completion 
is the task of predicting missing facts.
Here,
LLMs may either play the role of \emph{encoders} or \emph{generators}. 

When LLMs act as encoders~\cite{llm-as-encoders-choi-2021,llm-as-encoders-wang-2021,llm-as-encoders-shen-2023},
their goal is to encode textual information to ease the work of another model,
aimed at predicting missing facts.
%
%
When LLMs act as generators~\cite{kg-complenion-saxena-2022,kg-complenion-chen-2022,kg-complenion-xin-2022,kg-complenion-zhu-2023}
they extract the missing facts directly.
%
All such contributions differ from our work,
as they assume that information about the KG is somehow textually encoded into the LLM's prompt,
hence the LLM is a mere tool for extraction rather than the source from which to extract.
%
Technically speaking,
this implies that these methods require some input text from which to extract the missing facts,
while our work does not.

Most notably,
some methods
-- namely,~\cite{kg-complenion-chen-2022,kg-complenion-xin-2022} --
require the LLM to undergo ad-hoc training on a corpus of text data,
which is yet another requirement our work does \emph{not} have.
Unlike those works,
we do not require an additional pre-training phase for the LLM.
%

\paragraph{LLM-augmented KG construction}

KG construction is the task of building a KG from scratch.
Here,
LLM may assist in the process of entity discovery, end-to-end KG construction, or KG distillation.

Entity discovery aims at mining KG entities from textual data.
Some works~\cite{entity-discovery-ayoola-2022} use models to
first classify entities in text data, and then link them to form a KG.
%
Again,
this approach is different from ours,
as we do not rely on external text data.
Other methods~\cite{entity-discovery-decao-2021} train LLMs on a corpus of text data and use it to later retrieve entities.
These differ from our approach as we do not require any additional pre-training phase for the LLM.
%

End-to-end KG construction methods use LLMs to generate KGs in an end-to-end fashion.
Methods of this kind~\cite{end-to-end-kg-kumar-2020,end-tp-end-kg-melnyk-2021,end-to-end-kg-han-2023},
despite working differently,
still require ad-hoc textual data as the input for the LLM.
%
Therefore,
they differ from our work for the same reason as the above.

Finally,
there are a few works
-- namely, Comet~\cite{comet-2019} and Harvest~\cite{HaoTTNSZXH23} --
that perform KG distillation from LLMs,
meaning that they use LLMs as oracles to generate KGs,
similarly to our work.

In Comet~\cite{comet-2019},
an LLM is \emph{trained} to generate missing objects ($o$) in the triples of the form $(s, p, o)$.
The distillation procedure involves querying an LLM of such a sort.
While this task is similar to what we do in our method,
there are a few notable differences.
First,
Comet is \emph{not} tailored on ontologies,
meaning that it does not take into account the nature of $p$ and $s$ in distilling $o$,
which may lead to inconsistencies in the generated KG%
---a situation which is impossible by Construction in our method.
%
%
Second,
\cite{comet-2019} implies training an LLM in a particular way,
-- which is the main focus of that paper --
whereas our method requires no particular sort of training for the LLM,
aside general-purpose pre-training.

Another similar method is Harvest~\cite{HaoTTNSZXH23},
which relies on pre-trained LLMs to distill KGs.
The method generates novel triples of the form $(s, \mathsf{p}, o)$,
for given relations $\mathsf{p}$,
starting from prompts that exemplify those relations,
to be submitted to LLM oracles.
Peculiar to this method is the fact that prompt engineering is LLM-mediated too:
for each relation $\mathsf{p}$,
users just need to provide a natural language description and a few examples of the relation,
and further examples are generated by the LLM.
These examples are then used as prompts to query the LLM for pairs of entities $(s, o)$
which are related by $\mathsf{p}$.
Similarly to Comet and differently from our method,
Harvest is not tailored on ontologies,
so it may lead to inconsistent KGs too.
However,
Harvest, like our method, does not require any additional training phase for the LLM,
and can be applied to general-purpose LLMs.
Because of this similarity,
we provide a technical comparison between Harvest and our method in \Cref{ssec:comparison}.


\section{Filling Ontologies with \llmfkg{}}
\label{sec:contribution}

In this section,
we present \llmfkg{},
a framework for semi-automatic ontology population exploiting LLM as oracles.
We do our best to keep the description of \llmfkg{} abstract,
general and concise,
yet we acknowledge that,
to make the framework effective in practice,
it requires engineering several technical details,
which we discuss in \ref{ssec:practical-aspects}.

At the abstract level,
the core functioning of \llmfkg{} is very simple.
Stemming from a partially instantiated ontology including at least class and property definitions,
a set of query templates,
and a trained LLM oracle,
\llmfkg{} generates questions from the templates,
with the purpose of querying the oracle 
to populate the ontology from its answers.

\begin{figure*}
    \rotatebox{90}{
    \begin{minipage}{\textheight}
        \centering
        \includegraphics[width=\linewidth]{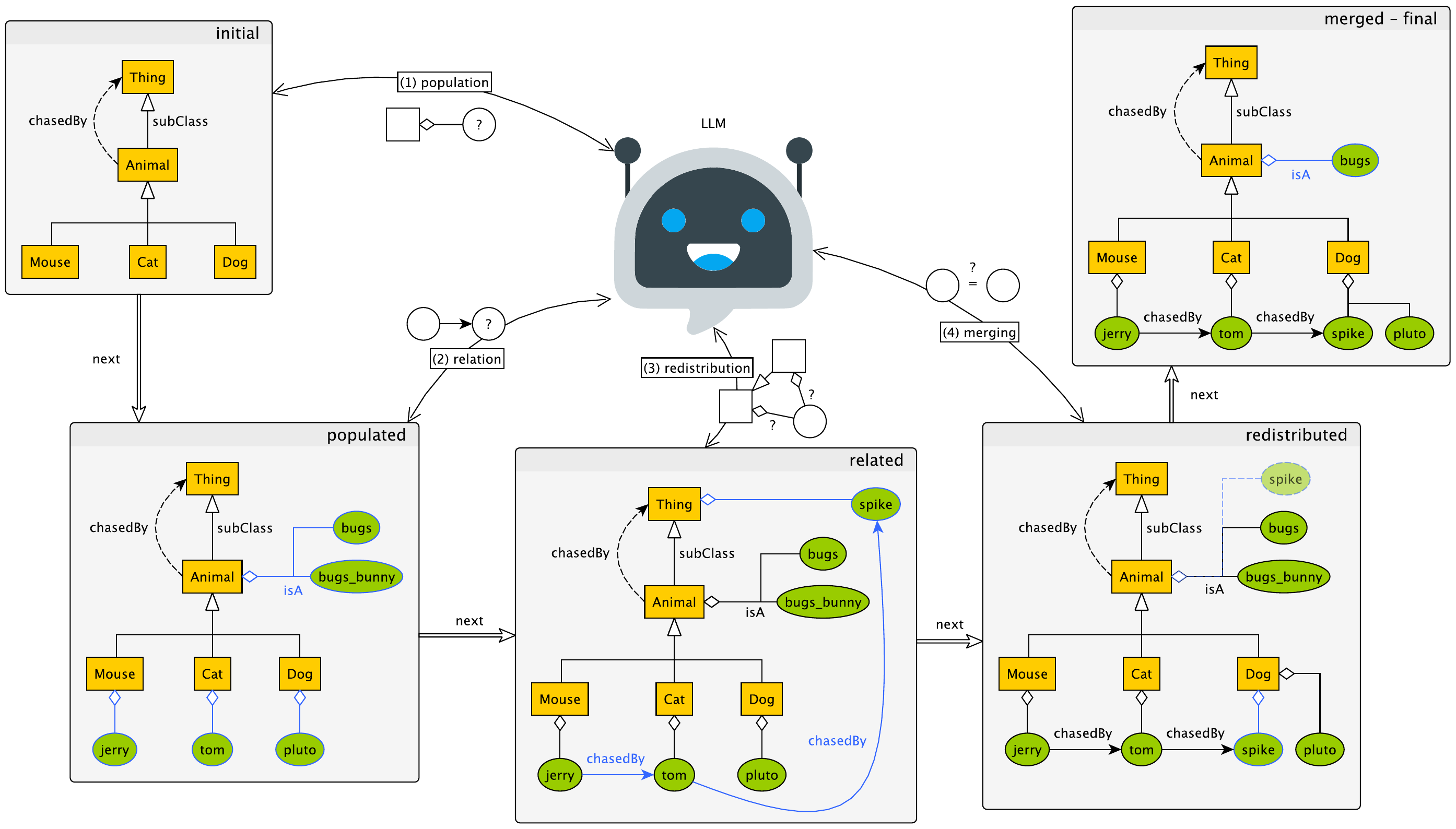}
        \caption{
            Overview of \llmfkg{}, based on a running example.
            The example assumes that the ontology to be populated is about animals,
            and it includes the classes $\concept{Cat}, \concept{Dog}, \concept{Mouse} \sqsubset \concept{Animal} \sqsubset \concept{Thing} \equiv \top$
            -- none of which has any instance yet --,
            as well as a property $\relation{chasedBy} : \concept{Animal} \times \top$,
            stating that each animal may be chased by some other entity.
            The initial state of the input ontology is depicted in the top-left box:
            classes are represented as yellow boxes,
            property definitions as dashed edges with black arrows,
            while subsumption relations among classes as straight edges with big white arrows.
            The \llmfkg{} algorithm will then encompass four phases,
            each one depicted in a separate box:
            in each box,
            differences with respect to the previous state are highlighted in blue.
            The bottom-left box represents the outcome of the first phase,
            namely the population phase,
            the LLM is queried to generate instances for the classes in the ontology.
            Instances are represented as green ellipses,
            whereas the relations between instances and classes are depicted as straight edges with white diamonds.
            Accordingly,
            in this phase,
            we let the LLM return animals from old cartoons such as:
            $\instance{tom}$ (which is a cat),
            $\instance{jerry}$ (which is a mouse),
            $\instance{pluto}$ (which is a dog),
            and $\instance{bugs}$ and $\instance{bugs\_bunny}$%
            ---i.e., two different names referencing the same entity
            (which is a rabbit).
            Rabbits get assigned to the $\concept{Animal}$ class,
            as no better class is available in the ontology.
            In the next phase (bottom-middle box),
            the relation phase,
            the LLM is queried to generate relations between instances, w.r.t. the properties in the ontology.
            The relations $\relation{chasedBy}(\instance{jerry}, \instance{tom})$ and $\relation{chasedBy}(\instance{tom}, \instance{spike})$,
            where $\instance{spike}$ novel instance of a dog,
            generated on the fly as in instance of $\top$%
            ---as $\top$ class is the range of $\relation{chasedBy}$.
            Such a new entry is eventually moved into the $\concept{Dog}$ class during the
            redistribution phase (bottom-right box),
            where the LLM is queried to redistribute instances among sub-classes.
            In the process,
            $\instance{spike}$ is moved from $\top$ to $\concept{Animal}$ and then to $\concept{Dog}$,
            where it belongs.
            Finally, in the merging phase (top-right box),
            the LLM is queried to merge instances that are syntactically similar.
            This is the case,
            for instance,
            of $\instance{bugs}$ and $\instance{bugs_bunny}$,
            which are merged into a single entity.
            This instance should remain in the $\concept{Animal}$ class,
            as no other class in the ontology is more specific than $\concept{Animal}$ for this particular individual.
        }
        \label{fig:roadmap}
    \end{minipage}}
\end{figure*}

An overview of the \llmfkg{} framework is depicted in \Cref{fig:roadmap},
leveraging on a simplified graphical representation of a run of the population algorithm.
While the rest of this section provides a detailed description of the framework,
the figure is intended as a compact roadmap to guide the reader through the various stages of the algorithm.

\paragraph{Input ontology}

\llmfkg{} assumes the ontology to be populated is already partially initialised.
%
The ontology must contain class and property definitions,
and classes must be organised hierarchically,
into a directed acyclic graph (DAG).
%
Classes may, or may not, have instances.
In other words,
the ontology may be partially populated with instances.

Classes must be named after the concepts they represent.
Class names must be meaningful, short, and unambiguous to avoid problems in later stages of the algorithm
(e.g., generation of queries for the LLM oracle).
A similar argument holds for the property names. 

Under such hypotheses,
\llmfkg{} will generate individuals for all the classes in the ontology,
and it will associate individuals by means of the properties in the ontology.
In doing so,
it will guarantee that generated individuals are represented by meaningful names,
and they are placed in the most specific class available in the ontology.

It is worth highlighting that
our ``partial initialisation'' assumption serves the use case where
ontology designers have already defined the concepts they want to represent in the ontology
and the relationships among them,
but they lack the instances to populate the ontology.
Conversely,
the ``partial population'' assumption serves the use case where
designers simply want to add ``more'' instances to the ontology.
The two use cases could be combined:
designers may define the skeleton of the ontology first
(i.e., the concepts and the relationships)
and they do several rounds of population via \llmfkg{},
possibly interleaved with manual interventions to refine the ontology.
In this scenario,
from the second round on,
\llmfkg{} would be applied to a partially instantiated ontology,
hence it must be designed to support this case.

\paragraph{Query templates}

\llmfkg{} needs to generate questions to be asked to the LLM oracle.
To do so,
it relies on a set of query templates,
which are pieces of text of arbitrary length (i.e. strings),
containing named placeholders.
Placeholders are meant to be eventually replaced with actual string,
by means of substitutions.
Substitutions are assignments of placeholders names to actual strings.
They can be applied to templates to generate concrete strings.

For example,
the template
$t$ = \str{What is the capital of \var{c}?}
may be instantiated by applying the substitution
$\sigma = \{\var{c}\mapsto\str{Italy}\}$,
denoted by $t / \sigma$.

We distinguish among query templates of four sorts: 
\begin{description}
    \item[individual seeking templates,] 
    asking for instances of a given class.
    %
    %
    For instance: 
    \str{examples of \var{class}?}.

    \item[relation seeking templates,]  
    asking for individuals related to some given individual via a given property.
    %
    %
    For instance: 
    \str{examples of \var{property} for \var{individual}?}.

    \item[best match templates,] 
    asking for the best class for a given individual,
    given a set of candidate classes.
    %
    %
    For instance: 
    \str{what class is best for \var{individual} among \var{classes}?}.

    \item[individuals merging templates,] 
    asking if two instances from the same class
    are semantically identical. 
    %
    %
    For instance 
    \str{in \var{class}, are the instances \var{ind_1} and  \var{ind_2} the same?}.
\end{description}

\paragraph{LLM oracle}

We do not impose any particular constraint on the nature of the LLM oracle,
nor on the sort of model it is based on.
In fact,
\llmfkg{} is agnostic to the LLM oracle,
and it can be used with any LLM oracle available on the market,
as far as it supports the generation of textual responses from textual prompts.
Both the prompt and the response are assumed to consist of strings of arbitrary length,
representing text in (some) natural language.
The natural language is assumed to be the same one used to name classes and properties in the ontology.
Without loss of generality, 
we assume the language is English.

Finally,
one key assumption is that the LLM oracle is already trained on a large corpus of textual data
-- also covering text in the natural language of choice --,
and that it absorbed a substantial amount of knowledge concerning the domain at hand.
%

\paragraph{Problem statement}

We formalise the ontology population problem solved by \llmfkg{} 
as follows.
Stemming from:
\begin{enumerate}
    \item a partially instantiated ontology $\relset{O} = \relset{C} \cup \relset{P} \cup \relset{X}$, including
    %
        a non-empty set of class definitions $\relset{C} \neq \varnothing$,
        a non-empty set of property definitions $\relset{P} \neq \varnothing$,
        and a possibly empty set of individuals and their relationships $\relset{X}$ (matching the aforementioned
        definitions);

    \item a subsumption relation $\sqsubseteq$ among the classes in $\relset{C}$;

    \item a set of query templates $\relset{T} = \relset{T}_{I} \cup \relset{T}_{R} \cup \relset{T}_{B} \cup \relset{T}_{M}$, where
    %
        $\relset{T}_{I}$ is a set of individual seeking templates,
        $\relset{T}_{R}$ is a set of relation seeking templates,
        $\relset{T}_{B}$ is a set of best match templates,
        and $\relset{T}_{M}$ is a set of individuals merging templates;

    \item and a trained LLM oracle $\relset{L}$, encapsulating domain-specific knowledge about $\relset{O}$;
\end{enumerate}
\llmfkg{} aims at generating a set of individuals and relationships $\relset{X}'$ such that
$\relset{X} \subseteq \relset{X}'$, and all novel individuals and relationships in $\relset{X}'$ are consistent with
the class and property definitions in $\relset{C}$ and $\relset{P}$, respectively.

In particular, \llmfkg{} guarantees that, at the end:
\begin{inlinelist}
    \item every individual is associated with the most specific concept available in $\relset{C}$ -- w.r.t. the
    subsumption relation $\sqsubseteq$ --,
    and
    \item for every property in $\relset{P}$, every individual in that property's domain is associated with several
    individuals in that property's range.
\end{inlinelist}
%

\paragraph{Phases}

To compute $\relset{X}'$, \llmfkg{} encompasses four major phases, namely:
\begin{inlinelist}
    \item the \textbf{population phase}, where novel individuals are identified for each class in $\relset{C}$,
    \item the \textbf{relation phase}, where novel relationships are identified for each property in $\relset{P}$,
    \item the \textbf{redistribution phase}, where the individuals identified in the previous phases are redistributed
    among the classes in $\relset{C}$, in such a way that each individual is put in the best (i.e. most specific) class
    available,
    \item the \textbf{merge phase}, where the individuals of each class in $\relset{C}$ are checked to detect
    semantic duplicates.
\end{inlinelist}
It is of paramount importance that phases follow this order of execution, as the population phase generates individuals
that are then used in the relation phase, and the relation phase may then generate novel individuals;
hence, the redistribution phase is needed to ensure that all individuals are put in the best class available,
while the merge phase is required to reduce the amount of semantically-duplicated entries.

\paragraph{Ancillary functions}

The remainder of this section relies on a number of ancillary functions,
whose details are provided in \ref{ssec:ancillary}.
Here,
we briefly summarise the notation.

Function \getRange{} (resp. \getDomain{}) returns the domain (resp. range) of a given property.
Function \askOracle{} models queries to LLM oracles,
hence it accepts and returns arbitrary strings.
Function \extractBinary{} (resp. \extractNames{}) aims to extract binary values
(resp. relevant individuals' or concepts' names) from LLMs' textual responses,
hence it returns a Boolean value (resp. a list of names),
while accepting a string as input.
Finally, function \addToClass{} adds an individual to a class,
doing nothing if the individual is already in the class,
or in any of its sub-classes.

\subsection{Population phase}

\begin{algorithm}
    \captionsetup{font=\algCaptionSize}
    \caption{Populates the given ontology with novel individuals queried from an LLM oracle}
    \label{alg:populate}
    \begin{algorithmic}[1]\algCodeSize
        \Require $\relset{O} = \relset{C} \cup \relset{P} \cup \relset{X}$: partially populated ontology
        \Require $\relset{T}_{I}$: individual seeking query templates
        \Require $\relset{L}$: LLM oracle
        \Require $\concept{R} \in \relset{C}$: root concept to be populated
        \Ensure $\relset{X}'$ contains novel individuals, assigned to the classes in $\relset{C}$
        \medskip
        \Function{\populate}{$\relset{O}, \relset{T}_I, \relset{L}, \concept{R}$}
            \State $\relset{X}' \gets \relset{X}$
            \ForAll{$\concept{C} \in \relset{C} \suchThat \concept{C} \sqsubset \concept{R} \wedge \concept{C} \neq \bot$}
                \State $\relset{O}' \gets \relset{C} \cup \relset{P} \cup \relset{X}'$
                \State $\relset{X}' \gets \Call{\populate}{\relset{O}', \relset{T}_{I}, \relset{L}, \concept{C}}$
            \EndFor
            \ForAll{$\mathit{t} \in \relset{T}_{I}$}
                \State $\mathit{q} \gets \mathit{t}/\{ \var{class} \mapsto \concept{R} \}$
                \State $\mathit{text} \gets \Call{\askOracle}{\relset{L}, \mathit{q}}$
                \ForAll{$\instance{i} \in \Call{\extractNames}{\mathit{text}}$}
                    \State $\relset{O}' \gets \relset{C} \cup \relset{P} \cup \relset{X}'$
                    \State $\relset{X}' \gets \Call{\addToClass}{\relset{O}', \instance{i}, \concept{R}}$
                \EndFor
            \EndFor
            \State \Return $\relset{X}'$
        \EndFunction
    \end{algorithmic}
\end{algorithm}

The population phase relies on the \populate{} function,
defined in \Cref{alg:populate}.
%
This function fills a partially instantiated ontology $\relset{O}$ with novel instances,
queried to some LLM oracle $\relset{L}$,
via a set of instance seeking query templates $\relset{T}_{I}$.

Starting from some root class of choice $\concept{R} \in \relset{O}$,
the function recursively explores the class graph spawned by the subsumption relation $\sqsubseteq$,
following a depth-first post-order exploration strategy.
%
%
%

For each visited sub-class $\concept{C} \sqsubseteq \concept{R}$,
the function generates as many queries as the templates in $\relset{T}_{I}$,
by replacing the placeholder $\var{class}$ in each template with the name of $\concept{C}$.
Then,
it submits each query to the LLM oracle $\relset{L}$,
and it extracts the names of the individuals from the response.
The amount of individual generated per query is unbounded,
and it depends on the LLM oracle $\relset{L}$,
and on other technicalities discussed in \ref{ssec:practical-aspects}.

It may happen that the same individual is generated by multiple queries,
or that the same individual is generated for
multiple different sub-classes of $\concept{R}$.
None of these situations is problematic,
as individual addition is performed by means of the \addToClass{} function,
which does not duplicate individuals,
nor assignments to classes,
and in case of multiple assignments,
prioritises the most specific class.
In keeping individuals assignment as specific as possible,
the \emph{post-order} exploration strategy is crucial.
By visiting most specific classes first,
the function ensures that the most specific concepts are populated first,
while less specific concepts are directly populated only when
no more specific concepts are available in the ontology.

Notably,
to populate the whole ontology $\relset{O}$,
one may simply invoke the function as follows:
\Call{\populate}{$\relset{O}, \relset{T}_I, \relset{L}, \top$}.
%

\subsection{Relation phase}
\label{sssec:relation}

\begin{algorithm}
    \captionsetup{font=\algCaptionSize}
    \caption{Populates the given ontology with novel relationships queried from an LLM oracle}
    \label{alg:relate}
    \begin{algorithmic}[1]\algCodeSize
        \Require $\relset{O} = \relset{C} \cup \relset{P} \cup \relset{X}$: partially populated ontology
        \Require $\relset{T}_{R}$: relation seeking query templates
        \Require $\relset{L}$: LLM oracle
        \Require $\relation{P} \in \relset{P}$: property to be populated
        \Ensure $\relset{X}'$ contains novel relationships, involving individuals in $\relset{X}$
        \medskip
        \Function{\relate}{$\relset{O}, \relset{T}_R, \relset{L}, \relation{P}$}
            \State $\relset{X}' \gets \relset{X}$
            \State $\concept{D} \gets \Call{\getDomain}{\relation{P}}$
            \State $\concept{R} \gets \Call{\getRange}{\relation{P}}$
            \ForAll{$(\instance{i} : \concept{D}) \in \relset{X}$}
                \ForAll{$\mathit{t} \in \relset{T}_{R}$}
                    \State $\mathit{q} \gets \mathit{t}/\{ \var{individual} \!\mapsto\! \instance{i}, \var{property} \!\mapsto\! \relation{P} \}$
                    \State $\mathit{text} \gets \Call{\askOracle}{\relset{L}, \mathit{q}}$
                    \ForAll{$\instance{i}' \in \Call{\extractNames}{\mathit{text}}$}
                        \State $\relset{O}' \gets \relset{C} \cup \relset{P} \cup \relset{X}'$
                        \State $\relset{X}' \gets \Call{\addToClass}{\relset{O}', \instance{i}', \concept{R}}$
                        \State $\relset{X}' \gets \relset{X}' \cup \{ \relation{P}(\instance{i}, \instance{i}') \}$
                    \EndFor
                \EndFor
            \EndFor
            \State \Return $\relset{X}'$
        \EndFunction
    \end{algorithmic}
\end{algorithm}

The relation phase relies on the \relate{} function,
defined in \cref{alg:relate}.
%
This function fills a partially instantiated ontology $\relset{O}$ with
novel relationships between the individuals therein contained
(and, possibly, novel individuals as well),
queried to some LLM oracle $\relset{L}$,
via a set of relation seeking query templates $\relset{T}_{R}$.

Focussing upon the some property $\relation{p} \in \relset{O}$
s.t. $\relation{p} : \concept{D} \times \concept{R}$,
%
the function
queries the LLM oracle $\relset{L}$ about relationships linking
each individual $\instance{i}$ in $\concept{D}$,
to some other individual $\instance{i}'$ in $\concept{R}$,
through $\relation{p}$.
%
In particular,
for each individual,
and for each query template $t \in \relset{T}_{R}$,
the function produces a query $q$
by replacing the placeholders $\var{property}$ and $\var{individual}$ in $t$
with the names of $\relation{p}$ and $\instance{i}$,
respectively.
Each query $q$ is then submitted to the LLM oracle $\relset{L}$,
and the names of the individuals in the response
are considered as individuals to be added to the ontology%
---and,
in particular,
to the range $\concept{R}$ of $\relation{p}$.
Accordingly,
for each individual $\instance{i}'$ returned by the LLM oracle,
the function shall add both the individual $\instance{i}'$
and the relationship $\relation{p}(\instance{i}, \instance{i}')$
to the ontology.

To populate relationships for all properties $\relation{p} \in \relset{O}$ in the ontology,
one may simply invoke the function once per property,
as follows:
\Call{\relate}{$\relset{O}, \relset{T}_R, \relset{L}, \relation{p}$}.

It is worth highlighting that the \relate{} function may generate novel individuals,
as a by-product of its operation.
When this is the case,
%
it may happen that the range $\concept{R}$ of the target relation $\relation{p}$,
is not necessarily the most adequate class for the generated individuals.
For instance,
there could be some sub-class $\concept{C} \sqsubset \concept{R}$ which is a better fit for some generated individuals.
This is why the redistribution phase is needed.

\subsection{Redistribution phase}
\label{sssec:redistribute}

\begin{algorithm}
    \captionsetup{font=\algCaptionSize}
    \caption{Redistributes individuals from the given ontology's classes, in such a way that each individual is assigned to the most specific class available}
    \label{alg:redistribute}
    \begin{algorithmic}[1]\algCodeSize
        \Require $\relset{O} = \relset{C} \cup \relset{P} \cup \relset{X}$: partially populated ontology
        \Require $\relset{T}_{B}$: relation seeking query templates
        \Require $\relset{L}$: LLM oracle
        \Require $\concept{R} \in \relset{C}$: root concept within which redistribution should occur
        \Ensure $\relset{X}'$ contains the different assignments of individuals to classes
        \medskip
        \Function{\redistribute}{$\relset{O}, \relset{T}_B, \relset{L}, \concept{R}$}
            \State $\relset{X}' \gets \relset{X}$
            \State $\relset{S} \gets \{ \concept{S} \in \relset{C} \mid \concept{S} \sqsubset \concept{R} \}$
            \ForAll{$(\instance{i} : \concept{R}) \in \relset{X}'$}
                \State $\concept{B} \gets \concept{R}$
                \ForAll{$\mathit{t} \in \relset{T}_{B}$}
                    \State $\mathit{q} \gets \mathit{t}/\{ \var{individual} \mapsto \instance{i}, \var{classes} \mapsto \relset{S} \}$
                    \State $\mathit{text} \gets \Call{\askOracle}{\relset{L}, \mathit{q}}$
                    \ForAll{$\concept{C} \in \Call{\extractNames}{\mathit{text}}$}
                        \State $\concept{B} \gets \concept{C}$
                        \State \Break{going to line \ref{line:first-result}}
                    \EndFor
                \EndFor
                \State $\relset{O}' \gets \relset{C} \cup \relset{P} \cup \relset{X}'$ \label{line:first-result}
                \State $\relset{X}' \gets \Call{\addToClass}{\relset{O}', \instance{i}, \concept{B}}$
            \EndFor
            \ForAll{$\concept{C} \in \relset{C} \suchThat \concept{C} \sqsubset \concept{R} \wedge \concept{C} \neq \bot$}
                \State $\relset{O}' \gets \relset{C} \cup \relset{P} \cup \relset{X}'$
                \State $\relset{X}' \gets \Call{\redistribute}{\relset{O}', \relset{T}_{B}, \relset{L}, \concept{C}}$
            \EndFor
            \State \Return $\relset{X}'$
        \EndFunction
    \end{algorithmic}
\end{algorithm}

The redistribution phase relies on the \redistribute{} function,
defined in \cref{alg:redistribute}.
%
This function 
redistributes individuals among the classes of an instantiated ontology $\relset{O}$,
in such a way that each individual is assigned to the most specific class available. 
%
To do so,
the function queries the LLM oracle $\relset{L}$ via a set of best match query templates $\relset{T}_{B}$.


Starting from some root class of choice $\concept{R} \in \relset{O}$,
the function recursively explores the class graph spawned by the subsumption relation $\sqsubseteq$,
following a depth-first \emph{pre}-order exploration strategy.
%
%
%
For each visited class $\concept{C} \sqsubseteq \concept{R}$,
the function attempts to determine whether $\concept{C}$ is actually the best class for all individual therein contained.
For some individuals,
it might be the case that $\concept{C}$ is not the best class,
as some of its sub-classes is more indicated.
%
For instance,
it may happen that classes $\concept{Cat} \sqsubset \concept{Animal}$ are available,
yet $\instance{tom}$ is assigned to $\concept{Animal}$ rather than to $\concept{Cat}$.

To determine whether $\concept{C}$ is the best class
(among it, and all its sub-classes)
for some individual $\instance{i} : \concept{C}$,
the function generates a query for each template $t \in \relset{T}_{B}$,
by replacing the placeholders $\var{individual}$ and $\var{classes}$ in $t$
with the name of $\instance{i}$ and
the concatenation of names of all direct sub-classes of $\concept{C}$,
respectively.
Each query $q$ is then submitted to the LLM oracle $\relset{L}$,
and the names of the individuals in the response
are considered as class names,
i.e. as candidate classes for $\instance{i}$.
Among these,
the function selects the first one
($B$, for ``best'')
and re-assigns $\instance{i}$ to $B$.

As the function is recursive,
each individual may be reassigned multiple times,
until eventually reaching the most specific class available for it in the ontology.
To make this possible,
the depth-first pre-order exploration strategy is crucial,
as it lets individuals ``move down'' in the class hierarchy quickly,
while keeping the size of queries relatively small.

Notably,
to redistribute \emph{all} individuals from an ontology $\relset{O}$,
one may simply invoke the
function as follows:
\Call{\redistribute}{$\relset{O}, \relset{T}_I, \relset{L}, \top$}. 

\subsection{Merge phase}\label{sssec:merge}

\begin{algorithm}
    \captionsetup{font=\algCaptionSize}
    \caption{Merges syntactically and semantically similar individuals from the given ontology's classes}
    \label{alg:merge}
    \begin{algorithmic}[1]\algCodeSize
        \Require $\relset{O} = \relset{C} \cup \relset{P} \cup \relset{X}$: partially populated ontology
        \Require $\relset{T}_{M}$: indivuals merging query templates
        \Require $\relset{L}$: LLM oracle
        \Require $\concept{R} \in \relset{C}$: root concept within which merge should occur
        \Ensure $\relset{X}'$ contains no duplicated individuals
        \medskip
        \Function{\merge}{$\relset{O}, \relset{T}_{M}, \relset{L}, \concept{R}$}
            \State $\relset{X}' \gets \relset{X}$
            \ForAll{$\concept{C} \in \relset{C} \suchThat \concept{C} \sqsubset \concept{R}$}
                \State $\mathcal{D} \gets \emptyset$
                \ForAll{$\instance{i}, \instance{j} : \concept{C} \suchThat ( \instance{i}, \instance{j} ) \in \relset{X}' \times \relset{X}'$}
                    \If{$\Call{\synSimilar}{\instance{i}, \instance{j}}$}
                        \State $\mathcal{D} \gets \mathcal{D} \cup \{\{\instance{i}, \instance{j}\}\}$
                    \EndIf
                \EndFor
                \ForAll{$\{ \instance{i}, \instance{j} \} \in \mathcal{D}$}
                    \ForAll{$\mathit{t} \in \relset{T}_{M}$}
                    \label{line:next-couple}
                        \State $\mathit{q} \gets \mathit{t}/\{ \var{ind_1}\!\mapsto\!\instance{i}, \var{ind_2}\!\mapsto\!\instance{j},
                            \var{class}\!\mapsto\!\concept{C} \}$
                        \State $\mathit{text} \gets \Call{\askOracle}{\relset{L}, \mathit{q}}$
                        \If{$\Call{\extractBinary}{text}$}
                            \State $\relset{X}' \gets \Call{\mergeInst}{i, j}$
                            \State \Break{going to line \ref{line:next-couple}}
                        \EndIf
                    \EndFor
                \EndFor
            \EndFor
        \EndFunction
    \end{algorithmic}
\end{algorithm}

The merge phase relies on the \merge{} function,
defined in \Cref{alg:merge}.
%
This function 
merges 
duplicated individuals from an instantiated ontology $\relset{O}$.

By `duplicated individuals',
we mean individuals with slightly different names,
which are actually semantically identical.
We observe empirically that these could appear in the ontology,
as a by-product of the previous phases.
Consider for instance the case: 
$\instance{maincoone\_cat}, \instance{maincoone} : \concept{Cat} \sqsubset \concept{Animal}$,
%
which are syntactically different.
To mitigate this issue,
the \merge{} function queries the LLM oracle to identify \emph{semantically} similar individuals,
and it merges them together into a single one.

To serve its purpose,
the \merge{} function scans an instantiated ontology $\relset{O}$,
looking for \emph{pairs} of syntactically similar individuals.
%
For each pair,
the function queries the LLM oracle $\relset{L}$
via a set of individuals merging templates $\relset{T}_{M}$,
to determine whether to merge them into a single individual or not.

More precisely,
starting from some root class of choice $\concept{R} \in \relset{O}$,
the function recursively explores the class graph spawned by the subsumption relation $\sqsubseteq$.
For each visited class $\concept{C} \sqsubseteq \concept{R}$,
the function first determines the set $\mathcal{D}$ of \emph{candidate} duplicate pairs,
by means of the \synSimilar{} function.
Then,
for each pair,
it decides whether to merge the two individuals or not,
by means of the \extractBinary{} function%
---which interprets the response from the LLM oracle as a boolean value.
In case of a positive response,
the \mergeInst{} function is exploited to perform the actual merge.
The latter simply works by transferring the information of an instance to the other
and by removing one of the two,
hence returning the updated ontology $\relset{X}'$.


As far as LLM queries are concerned,
these rely on the templates in $\relset{T}_{M}$.
For each pair of candidate duplicates,
the \merge{} function generates a query for each template $t \in \relset{T}_{M}$,
by replacing the placeholders $\var{ind\_1}$, $\var{ind\_2}$, and $\var{class}$ in $t$
with the name of the instances $\instance{i}$ and $\instance{j}$,
and of class $\concept{C}$,
respectively.
The first template producing a positive response is considered as definitive,
otherwise further templates are queried.
Most notably,
to avoid issues that may arise from the LLM's creativity,
we force-set temperature to 0.0 when querying the LLM oracle in the \merge{} function.
%

%

Notably,
to detect and merge \emph{all} duplicated individuals from an ontology $\relset{O}$,
one may simply invoke the function as follows:
\Call{\merge}{$\relset{O}, \relset{T}_M, \relset{L}, \top$}. 

\paragraph{Remarks}

Despite algorithm \ref{alg:merge} 
uses a post-order strategy,
the choice of the exploration strategy is \emph{not} relevant here.
Similar formulations may be written,
leveraging different strategies.
These would be equivalent,
as long as they visit all direct and indirect sub-classes of $\concept{R}$.

Another remark is about the role of the \synSimilar{} function.
As it decides if two instances
$\instance{i}$ and $\instance{j}$
should be considered \emph{enough} similar at the syntactic level
to be considered for merging,
%
%
the implementation of this function has a relevant impact on the overall performance of the \merge{} function,
as we further discuss in \ref{ssec:practical-aspects}.

Finally,
about the overall role of the merging phase,
one may wonder why we do not rely on entity alignment approaches \cite{ZhaoTkde2022}.
While these mechanisms are effective in identifying similar instances in different knowledge graphs,
they usually require additional data about the domain at hand \cite{ChenIjcai2018,XuAcl2019}.
This is because
entity alignment systems rely on ad-hoc ML models being trained on the alignment task,
thus introducing an additional layer of complexity in the system.
In fact,
in our approach,
we consider relying on the same LLM oracle for both populating the ontology and to identifying the duplicates in it.
While not completely error-free,
this approach avoids additional computational burdens in the population phase,
while prevents the need,
for the user,
to provide additional data about the domain at hand,
as well as to train additional models.

\section{Case Study and Experiments}
\label{sec:case-study}

In this section,
we present a case study we designed to validate \llmfkg{} empirically.

\paragraph{Experimental Setup}

The experimental setup is as follows.
We design a non-trivial ontology (cf.\ \Cref{ssec:ontology}),
by defining a class hierarchy and a set of properties,
and we populate it by means of \llmfkg{}.
In doing so,
we fine-tune ad-hoc query templates (cf.\ \Cref{ssec:query-templates}),
we exploit several LLM oracles,
from different families and technologies (cf.\ \Cref{ssec:llms-oracles}),
hence producing a set of populated ontologies,
all sharing the same class hierarchy and properties.
We then analyse and compare the populated ontologies,
with the purpose of assessing the performance of \llmfkg{},
and the impact of the LLM family of choice on the quality of the population process.
To do so,
we define a taxonomy of errors that may occur during the population process,
as well as a set of metrics (cf.\ \Cref{ssec:performance-metrics}) based on those errors,
to measure the quality of the populated ontologies.
We then \emph{manually} inspect the generated ontologies
(one-by-one and one individual at a time),
looking for errors and computing the corresponding scores accordingly
(cf.\ \Cref{ssec:results}).

Testing \llmfkg{} on a custom ontology created for the purpose of this case study
-- as opposed to testing it on a publicly available ontology --
comes with several key benefits.
First,
it allows us to finely control every aspect of the experiment%
---there including its complexity and its content.
Second,
it avoids the hard-to-verify and hard-to-exclude situation where
LLM oracles perform good simply because they are tested on some ontology they have already met during training.
However,
it also comes with some drawbacks%
---most notably,
the fact that no ground truth data is available for the ontology,
hence no simple automatic validation procedure can be devised.
This is the reason why we resort to \emph{manual} inspection of the populated ontologies for validation:
thoroughly inspecting each edit made by \llmfkg{} to the ontology is the only way to spot errors
and check that the output of our algorithm is correct.

\paragraph{Code and Data}

Technically speaking,
we conduct our experiments by means of a Python implementation of \llmfkg{},
developed by us,
which is publicly available on GitHub\footnotemark.
\footnotetext{\url{https://github.com/Chistera4-Expectation/kg-filler}}
The experimental results are available as well on GitHub,
on a dedicated repository\footnotemark.
\footnotetext{\url{https://github.com/Chistera4-Expectation/knowledge-graphs/branches}}
There,
each Git branch corresponds to a different experiment,
whereas the \texttt{main} branch contains the initial ontology%
---only containing class and property definitions,
and no individuals.

\paragraph{Reproducibility}

For the sake of reproducibility,
each experiment branch on GitHub contains not only the ontology populated by \llmfkg{} in that experiment,
but also the caches of the LLM query--response pairs,
for all the queries submitted to the LLM oracle during the experiment.
In this way,
anyone can inspect the exact queries submitted to the LLM oracle,
and the exact responses received,
and can reproduce our experiments deterministically.

Finally,
we automate our experiments so that each experiment's branch contains
a sequence of commits such that every commit corresponds to
a unitary operation that \llmfkg{} performs on the ontology
(e.g., populating a class with instances, populating a property with relationships, etc.).
In this way,
the history of commits in each branch can be used to inspect the exact sequence of operations
performed by \llmfkg{} in the corresponding experiment.

\subsection{Ontology}
\label{ssec:ontology}

As dictated by the needs of the \expectation{} project~\cite{expectation-extraamas2021},
(i.e., one of the funding sources of this work),
we design an ontology in the \emph{nutritional} domain.
In \expectation{},
the ontology would serve as a basis for a nutritional recommender system,
aimed at suggesting recipes to users based on their dietary needs and preferences
(cf.\ \cite{skerecommender-cmbp235}).

Despite such recommender system lays outside the scope of this work,
we believe this detail may help the reader in understanding the ontology design choices.
We also stress that,
as far as this work is concerned,
the ontology is just one of many possible case studies,
and it was designed by only leveraging on the authors' common-sense.
Designing a scientifically accurate nutritional ontology would require the expertise of a nutritionist,
and this lays outside the scope of this work.

The ontology aims at collecting \emph{recipes} and their \emph{ingredients}.
To do so,
it categorises \emph{edible} items into several sub-classes,
which should allow a recommender system to categorise edible items
in such a way to match all possible dietary needs and preferences.

\begin{figure*}
    \centering
    \includegraphics[width=\linewidth]{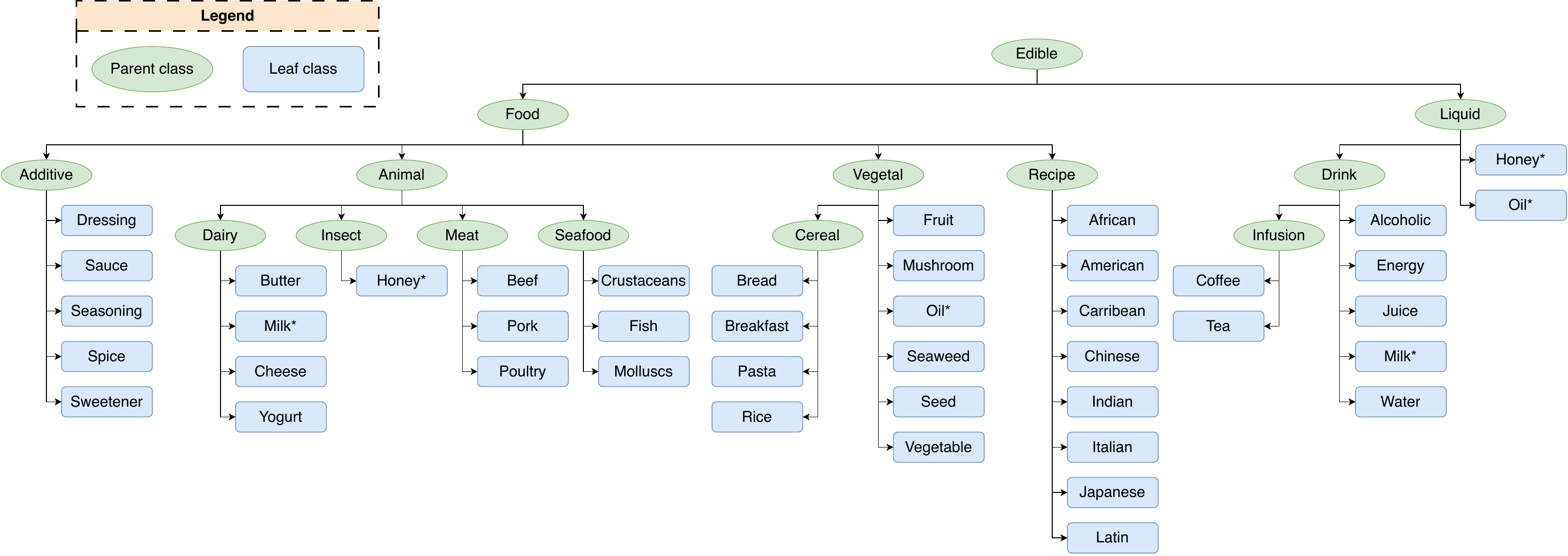}
    \caption{
        Class hierarchy of the case study ontology.
        Notice that the hierarchy is not really a tree, but rather a DAG.
        The asterisk (*) denotes classes having multiple super-classes
        (they are depicted once per super-class for the sake of readability).
    }
    \label{fig:onto}
\end{figure*}

The ontology consists of \numberOfClasses{} classes,
 organised as in \cref{fig:onto}.
The root class is \concept{Edible} and it has
several direct or indirect subclasses.
One relevant sub-class of \concept{Edible} is \concept{Recipe},
which is the root class of various sorts of cuisines
(e.g.\ \concept{Italian}, \concept{Chinese}, etc.).
%
%
Some classes have an annotation property,
(\relation{fancyName}) that better describes the meaning of the class
(cf.\ \ref{par:fancyname}).
%
%
Another relevant property is $\relation{ingredientOf} : \concept{Edible} \times \concept{Recipe}$,
which is used to connect recipes to their edible ingredients.
Of course,
the ontology is just a skeleton,
and it initially contains no individuals.

%

\subsection{Query Templates}
\label{ssec:query-templates}

To query the LLM oracle,
we fine-tune query templates as follows.

We define the set of individual seeking templates ($\relset{T}_{I}$),
as follows\footnotemark:
\str{$ ( $instances$ \mid $examples$ ) $ list for class \var{class}$ ( $, names only$ )? $}
\footnotetext{
    we denote by $(A|B)$ a template coming in two variants: one with $A$ and one with $B$;
    and by $(A)?$ a template with two variants: one with $A$ and one without it.
}

We define the set of relation seeking templates ($\relset{T}_{R}$)
as the singleton: \str{ingredient list for \var{individual}, names only}.
We only use a template with hard-coded property name,
to make the sentence look more natural.

We define the set of best match templates ($\relset{T}_{B}$)
as the singleton: \str{most adequate class for \var{individual} among: \var{classes}. concise}.

Finally,
we define the set of individual merging templates ($\relset{T}_{M}$)
as the singleton: \str{in the \var{class} class, should instances \var{ind_1} and \var{ind_2}
be merged together as semantic and ontologic duplicates? yes or no answer only}.

\subsection{LLM Oracles}\label{sssec:llms}
\label{ssec:llms-oracles}

To test the impact of the LLM quality on the \llmfkg{} ontology population process,
we consider integrating several different state-of-the-art LLMs into the \llmfkg{} pipeline.
In particular,
we consider 8 different LLMs aiming at covering the almost totality of the state-of-the-art.
We consider both open source
(e.g., OpenChat, Llama, etc.)
and closed models
(e.g., GPT),
as well as Mixture of Experts (MoE) models%
---e.g., Mixtral.

To integrate querying open source models into the \llmfkg{} pipeline,
we rely on the Hugging Chat API\footnote{\url{https://github.com/Soulter/hugging-chat-api}},
which represents a third party library of the Hugging chat service built by the HuggingFace team\footnotemark.
\footnotetext{\url{https://huggingface.co}}
Leveraging the Hugging Chat API it is possible to define multiple simultaneous conversations with different LLMs
hosted by the Hugging Chat platform\footnote{\url{https://huggingface.co/chat/}}.
Most popular HuggingFace open-source LLMs can be seemingly integrated into \llmfkg{},
and automatically queried via the Hugging Chat API\footnote{\url{https://github.com/Soulter/hugging-chat-api}}.
Therefore,
we select the following open source models for our evaluation:
\begin{description}
    \item[Llama 2:] a family of open-source LLMs,
    representing the most popular open-source alternative to OpenAI's GPT \cite{llama2-2023}.
    Throughout our experiments,
    we select the 70 billion parameters model of this family,
    as supported by the Hugging Chat API at the time of writing.

    \item[OpenChat 3.5:] a 13 billion parameters model fine-tuned from the Llama LLM family,
    optimised for mixed-quality data \cite{wang2023openchat}.

    \item[Mistral 7B:] a relatively small LLM
    (7 billion parameters)
    developed by the Mistral AI team\footnote{\url{https://mistral.ai}}
    and sponsored as the most powerful language model for its size \cite{mistral}.

    \item[Gemma:] a family of lightweight,
    state-of-the-art open models from Google,
    built from the same research and technology used to create the Gemini models \cite{gemini}.
    Throughout our experiments we select the 7 billion parameters model. 
\end{description}
Similarly,
we rely on the same Hugging Chat API and platform to include a few MoE models in our analysis.
MoE models combine multiple LLMs
-- possibly optimised on different data and of different sizes --
using ensemble techniques like boosting and bagging to achieve higher generalisation and reasoning performance.
In particular,
we consider the following MoE LLM models:
\begin{description}
    \item[Mixtral:] a pretrained generative sparse MoE
    -- developed by the Mistral AI team\footnote{\url{https://mistral.ai}} --
    aggregating 8 different LLMs, each having 7 billion parameters \cite{mixtral}.

    \item[Nous Hermes:] the new flagship Nous Research\footnote{\url{https://nousresearch.com}} model,
    trained over GPT-4 generated data
    -- as well as other high quality data --,
    achieving state-of-the-art performance on a variety of tasks.
    Similarly to Mixtral,
    this MoE model is built from the ensemble of 8 different 7 billion parameters LLMs.
\end{description}
Finally,
concerning closed source models,
we consider to rely on the OpenAI API\footnote{\url{https://openai.com/blog/openai-api}} to query two different versions of the GPT family of models, namely:
\begin{description}
    \item[GPT 3.5 Turbo:] the most popular LLM,
    showcasing ground-breaking performances against its predecessors and counterparts \cite{gpt3-2020}.

    \item[GPT 4 Turbo:] the latest version of the GPT models family \cite{gpt4}.
    Its exact number of parameters is unknown,
    but few sources report the GPT 4 model to be a MoE.
\end{description}
The LLM size is well-known to impact greatly the performance of the model,
thus affecting the overall quality of the pipeline relying on the LLM.
Similarly,
the experiments setup may alter the performance of the different models selected.
Therefore,
we report in \Cref{tab:llms_sizes} the LLM number of parameters along with few relevant hyperparameters of our experiments.
Here,
\emph{max tokens} represents limit of the LLM's output length,
while \emph{max retries} is used to limit the number of times the API is called in case of failure.
The \emph{back-off time} parameter is used to specify the time to wait between two consecutive API calls in case of failure.
Finally,
as LLMs are proven to be influenced by their \emph{temperature} parameter
which represents a cumbersome value to exactly fine-tune,
we 
leave it untouched to its default value
-- available either through the HuggingFace or the OpenAI API --
throughout all our experiments.
\Cref{tab:llms_sizes} highlights a relevant difference in terms of model size between open-source and close-source LLMs,
with GPT 4 being almost 200 times larger than the largest open source model available (Llama 2).
We expect this difference to profoundly impact the quality of the constructed ontology and delve more into this issue in \Cref{ssec:results}.

\begin{table}
    \caption{
        Size (number of parameters) and experiments setup of the different LLMs used in \llmfkg{}.
        Values reported with $^{\ast}$ are an educated estimate
        -- not confirmed --,
        as the corresponding models have not be fully disclosed to the public.
    }
    \resizebox{\columnwidth}{!}{
    \centering
    \begin{tabular}{l | c c c c c} \toprule
        LLM & size [B] & max tokens & max retries & back-off time [s] & temperature \\ \midrule
        GPT 3.5 Turbo \cite{gpt3-2020} & 375$^{\ast}$ & 1000 & 2 & 30 & 0.7 \\
        GPT 4 Turbo \cite{gpt4} & 1500$^{\ast}$ & 1000 & 2 & 30 & 0.7 \\ \midrule
        Openchat \cite{wang2023openchat} & 13 & 1000 & 2 & 30 & 0.1 \\
        Llama2 \cite{llama2-2023} & 70 & 1000 & 2 & 30 & 0.1 \\
        Mistral \cite{mistral} & 7 & 1000 & 2 & 30 & 0.1 \\
        Gemma \cite{gemini} & 7 & 1000 & 2 & 30 & 0.1 \\ \midrule
        Mixtral \cite{mixtral} & 56 & 1000 & 2 & 30 & 0.1 \\
        Nous Hermes & 56 & 1000 & 2 & 30 & 0.1 \\ \bottomrule
    \end{tabular}
    }
    \label{tab:llms_sizes}
\end{table}

\subsection{Performance Metrics}
\label{ssec:performance-metrics}

Measuring the quality of ontology construction and population processes is not trivial,
as several components come into play,
such as the quality of the added instances,
the reliability of the relations or their lack there-of.
Therefore,
we here 
analyse and present the possible issues that we should take into account when analysing the \llmfkg{}'s outcome,
and we define the corresponding quantitative performance metrics.

\subsubsection{Types of Errors}
\label{sssec:error_types}

When relying on sub-symbolic black-box models for populating ontologies,
several issues may arise depending on the quality of the oracle's output
and the quality of the output post-processing procedure.
LLMs are famously prone to hallucinations,
thus making their suggestions often unreliable.
Moreover,
their output's structure is often unpredictable,
depending on how the model interprets the given query,
thus the post-processing procedure is complex and not always infallible.
Therefore,
we here define several types of errors that may arise in the ontology population process.

\paragraph{Misplacement error ($E_{mis}$)}

The individual is part of the ontology,
but it is assigned to the wrong class.
This error represents the most common type of issue when dealing with large ontologies with many classes.
We distinguish between two sorts of misplacement errors,
depending on whether the misplaced individual is assigned to
\begin{inlinelist}
    \item a too general class -- i.e., there exists a better-suited sub-class for it --;
    or
    \item a semantically wrong class.
\end{inlinelist}
Throughout our investigation,
we consider these errors as quantitatively identical,
meaning that the occurrence of a misplacement is counted independently of the misplacement type.
However,
we argue that the first sort of misplacement error 
is `just' lack of precision,
while the second sort is a more severe issue.

\paragraph{Incorrect individual ($E_{ii}$)}

The individual makes no sense for the ontology,
yet it has a meaningful name.
This error occurs either when
\begin{inlinelist}
    \item the LLM oracle model produces an hallucinated response to a query,
    suggesting that an out-of-scope meaningful instance should be added to the ontology
    -- e.g., suggesting that ``\emph{dog}'' represents an instance of the class ``\emph{Edible meat}'' --;
    or
    \item the LLM output processing procedure fails,
    extracting a meaningful, yet out-of-context, instance from a correct LLM suggestion%
    ---e.g., wrongly post-processing the suggestion ``\emph{Bolognese sauce}''
    may result in adding the instance ``\emph{Bolognese}''
    to the class ``\emph{Recipe}'',
    while the word alone indicates a person from Bologna (Italy).
\end{inlinelist}
As these errors are generally related to LLM's hallucination issues,
they represent a relevant class of errors which require careful consideration.

\paragraph{Meaningless individual ($E_{mi}$)}

The individual name makes no sense at all,
and it should not be considered as a part of the ontology.
These errors occur when the LLM output processing procedure fails to identify a negative response 
or it fails to identify the correct cluster of words that refer to the suggested instance.
For example,
failing to interpret the message
``\emph{As an AI language model I do not have access to a pre-existing list of instances for class milk that contains names only}''
as a negative reply would create a meaningless instances in the ontology.

\paragraph{Class-like individuals ($E_{ci}$)}
The individual has the same or very similar semantic value of the class it belongs.
Broad answers from the LLM oracle cause this issue,
translating to vague instances that should be removed from a class.
For example,
when querying for ``\emph{meat}'' instances,
it is not uncommon for LLM oracles to suggest the same ``\emph{meat}'' to be part of the class itself.

\paragraph{Duplicate individuals ($E_{di}$)}

Two or more individuals have the same or very similar semantic value.
This issue may occur either from
\begin{inlinelist}
    \item the LLM oracle suggesting multiple times very similar instances
    -- e.g., apple, apples, apple slices, etc. --
    or from
    \item a wrongful post-processing of the LLM outputs.
\end{inlinelist}
Similarly,
as the proposed ontology filling process is iterative,
it is possible that different queries outputs produce semantically identical instances to be added to the ontology.
In \llmfkg{},
we aim at overcoming such issue with the latest \emph{merge phase} (cf.\ \Cref{sssec:merge}),
where we sample possible duplicates amongst the ontology class and query the LLM to assess if they need to be merged or not.
However,
this process is prone to error and duplicate individuals may still be present in the final ontology.

\paragraph{Wrong relation ($E_{wr}$)}

The relation linking two individuals is not valid.
In \llmfkg{}'s relation phase (\Cref{sssec:relation}),
we aim at populating a partially instantiated ontology $\relset{O}$
with novel relationships between the individuals therein contained.
This process is performed via querying the LLM oracle.
Therefore,
the relationships contained in the ontology may suffer issues related to the quality of the LLM's output
-- e.g., hallucinations --,
translating to wrong or invalid relationships between individuals.
For example,
an hallucinated LLM may suggest that ``\emph{onions}'' are an ingredient of the recipe ``\emph{carbonara spaghetti}'',
thus generating an inconsistent relation in the ontology.
Measuring these errors is fundamental to identify the quality of the generated ontology and connect it to the quality of the LLM's suggestions.

\paragraph*{About subjectivity}
Throughout our experiments,
we measure the impact of these errors by simply counting how many times they occur.
While the definition of these errors is arguable clear and objective,
the identification of the errors themselves is open to a small degree of subjectivity.
This is particularly the case for the `incorrect individual' and the `wrong relation' errors,
where the classification of an individual or a relation as `incorrect' or `wrong' may depend on the evaluator's interpretation.
To mitigate this issue,
we let each author evaluate the populated ontologies independently,
and we resolve any disagreement by majority voting.

\subsubsection{Measures}
\label{sssec:measures}

As \llmfkg{} relies on sub-symbolic models to populate the given ontology,
we need to assess the quality of the population process.
However,
up to our knowledge there exist no well-established quantitative measures that analyse the quality of a constructed ontology.
Therefore,
we define the following metrics,
focusing on various aspects of the individuals and relations quality:
\begin{enumerate}
    \item the total amount of generated individuals $\mathit{TI}$,
    so as to evaluate \llmfkg{}'s ability to generate large ontologies with various individuals for each class

    \item the minimum (resp. maximum) class weight $\mathit{minCW}$ (resp. $\mathit{maxCW}$),
    which is defined as the minimum (resp. maximum) amount of individuals generated per class.
    This measure allows for checking if
    -- and to what extent --
    the populated classes are unbalanced

    \item the total number of individuals belonging to leaf classes $\mathit{TL}$,
    gauging how much \llmfkg{} can identify specific classes for the generated individuals

    \item the total amount of individuals affected by errors $\mathit{TE}$%
    ---where each individual can count as either zero or one error

    \item the relative individual error $\mathit{RIE}$,
    defined as the ratio between the number of individuals affected by errors and the total number of individuals
    $\mathit{RIE} = \dfrac{\mathit{TE}}{\mathit{TI}}$

    \item the total amount of errors for each class of errors identified in \Cref{sssec:error_types}%
    ---each individual may count for zero, one, or more sorts of errors

    \item the total amount of generated relations between individuals $\mathit{TR}$

    \item the relative relation error $\mathit{RRE}$,
    which is simply defined as the ratio between the number of wrongful relation and the total number of relations
    $\mathit{RRE} = \dfrac{E_{wr}}{\mathit{TR}}$.
\end{enumerate}
The selected metrics used to measure the quality of the populated ontology
-- along with the corresponding types of errors presented in \Cref{sssec:error_types} --
were manually evaluated for each experiment considered in this paper.
%
%

Finally,
we define a quality metric $Q$ which summarizes the overall quality of the populated ontology
by measuring the correct individuals and relations over the total amount of generated instances.
More formally:
\begin{equation}
    Q = \dfrac{TI-TE + TR - E_{wr}}{TI + TR}.
\end{equation}
The quality metric $Q \in \left[0,1\right]$ represents an intuitive and easy-to-understand measure of the proposed automatic ontology population process,
being equal to $1$ for a perfect population mechanism that generates only valid instances and relations amongst them,
while being equal to $0$ whenever the LLM fills the ontology only with invalid data.

\subsubsection{Quality of Service}

Finally,
for the sake of comparison,
we consider three more metrics,
which aim at measuring the quality of service (QoS) of the LLM oracle
backing our \llmfkg{} algorithm.
These metrics are:
\begin{itemize}
    \item the time ($\mathit{\Delta t}$) required to populate the ontology
    (i.e., the time required by a single \llmfkg{} run),
    computed as the difference between the time of the last and the first query to the LLM oracle;

    \item the number of queries ($\mathit{N}$) submitted to the LLM oracle \emph{in total}
    by a single \llmfkg{} run
    (there including inconclusive queries);

    \item the total cost ($\$$), in USD, of a single \llmfkg{} run
    (considering that queries are cached on a per-model basis,
    so that the unitary cost of each query is computed at most once).
\end{itemize}

Care should be taken in interpreting these metrics,
because:
\begin{inlinelist}
    \item the time is impacted by network latencies, network congestion,
    and the LLM's service provider's load and rate limitations;

    \item the overall amount of queries is influenced by the length of the LLM's responses
    -- longer responses in earlier phases shall produce more queries in later phases --,
    as well as by the response length limitations imposed to save money;

    \item the cost is influenced by the number of queries,
    by their prompt,
    by the verbosity of the LLM's responses,
    by the model,
    and ultimately by the LLM's service provider's price.
\end{inlinelist}
Also notice that,
at the time of writing,
Hugging Face models can be queried for free
(with strict rate limitations),
while OpenAI ones come with the following pricing scheme
(with much looser rate limitations):
\begin{description}
    \item[GPT 3.5 Turbo:] \$0.5 (resp. \$1.5) every 1M \emph{input} (resp. \emph{output}) tokens;
    \item[GPT 4 Turbo:] \$10 (resp. \$30) every 1M \emph{input} (resp. \emph{output}) tokens.
\end{description}
Details about how \llmfkg{} deals with these aspects are provided in \ref{ssec:costs-rates}.

\subsection{Results}
\label{ssec:results}

\begin{table*}
    \caption{
        Performance of \llmfkg{} over different state-of-the-art LLMs.
        For each performance metric (column) we denote with $^{\dag}$ and $^{\ddag}$
        the best and second best performing model respectively.
    }
    \resizebox{\textwidth}{!}{
    \centering
    \begin{tabular}{l | c c c c c c c c c c c c c c | c} \toprule
        LLM & $\mathit{TI}$ & $\mathit{minCW}$ & $\mathit{maxCW}$ & $\mathit{TL}$ & $\mathit{TE}$ & $\mathit{RIE}$ & $E_{mis}$ & $E_{ii}$ & $E_{mi}$ & $E_{ci}$ & $E_{di}$ & $\mathit{TR}$ & $E_{wr}$ & $\mathit{RRE}$ & $Q$
        \\ \midrule
        GPT 3.5 Turbo \cite{gpt3-2020} & 511 & 4 & 40 & 495 & 44$^{\dag}$ & 0.0861$^{\dag}$ & 37$^{\dag}$ & 1$^{\dag}$ & 0$^{\dag}$ & 13$^{\dag}$ & 3$^{\dag}$ & 736 & 51$^{\dag}$ & 0.0693$^{\dag}$ & 0.924$^{\dag}$
        \\
        GPT 4 Turbo \cite{gpt4} & 735 & 5 & 56 & 727 & 91$^{\ddag}$ & 0.1238 & 40$^{\ddag}$ & 5 & 18 & 19 & 8 & 1061 & 96$^{\ddag}$ & 0.0905$^{\ddag}$ & 0.896$^{\ddag}$ 
        \\ \midrule
        Openchat \cite{wang2023openchat} & 997 & 9 & 105 & 970 & 244 & 0.2447 & 105 & 65 & 11 & 23 & 32 & 1329$^{\dag}$ & 440 & 0.3311 & 0.706 
        \\
        Llama2 \cite{llama2-2023} & 665 & 6 & 55 & 609 & 119 & 0.1789 & 84 & 8 & 4$^{\ddag}$ & 21 & 7 & 1062 & 197 & 0.1855 & 0.817 
        \\
        Mistral \cite{mistral} & 1176$^{\dag}$ & 11$^{\dag}$ & 110 & 1113$^{\dag}$ & 255 & 0.2168 & 133 & 14 & 60 & 32 & 72 & 1211 & 284 & 0.2345 & 0.774 
        \\
        Gemma \cite{gemini} & 433 & 1 & 40 & 407 & 107 & 0.2471 & 70 & 2$^{\ddag}$ & 21 & 16$^{\ddag}$ & 3$^{\dag}$ & 989 & 290 & 0.2932 & 0.721 
        \\ \midrule
        Mixtral \cite{mixtral} & 841 & 3 & 111$^{\ddag}$ & 819 & 228 & 0.2711 & 94 & 22 & 78 & 22 & 15 & 960 & 174 & 0.1813 & 0.777 
        \\
        Nous Hermes & 1121$^{\ddag}$ & 11$^{\dag}$ & 121$^{\dag}$ & 1082$^{\ddag}$ & 123 & 0.1097$^{\ddag}$ & 59 & 12 & 11 & 28 & 11 & 1222$^{\ddag}$ & 227 & 0.1858 & 0.851 
        \\ \bottomrule
    \end{tabular}
    }
    \label{tab:performance}
\end{table*}

\Cref{tab:performance} presents the results of the performances obtained by different runs of \llmfkg{},
when backed by the many LLM models considered in \Cref{sssec:llms}.
Overall,
the obtained results highlight how it is possible to build quite reliably ontologies
from the implicit knowledge that LLMs attain during their training process.
In fact,
focusing on the quality metric $Q$,
at least 75\% of the instances and relations added to the initial empty ontology are valid for most of the given LLMs.
Moreover,
GPT 3.5 Turbo achieves 92.4\% of valid instances and relations
and its successor GPT 4 Turbo achieves similar performance
-- i.e., $Q=89.6\%$ --,
encouragingly showing the reliability of our approach.

The types of errors occurred in the ontology population phase also represents a valuable insight,
as they highlight how most issues are related to the relative position of the added instances,
rather than highlighting the prevalence of nonsensical entries.
These results represent an encouraging finding
as they confirm the feasibility of relying on large general-knowledge black-box LLMs to extract
a structured and precise definition of the knowledge concerning a specific topic%
---such as the food domain in our case study.

Depending on the specific performance metric considered,
different models achieve the highest performance.
For example,
\emph{Mistral} constructs one of the largest ontology,
being capable of adding 1176 instances and 1211 relations to the final version of the ontology.
However,
the large number of instances and relations brings issues concerning the quality of the added components,
resulting in a high relative error measurement.
On the other hand,
GPT-based solutions achieve low relative error metrics,
with \emph{GPT 3.5 Turbo} achieving an encouraging $0.0861$ relative individual error
and an even more impressive relative relation error of $0.0693$.
However,
relying on these models result in the construction of relatively small ontologies,
as the \emph{GPT 3.5 Turbo} version has almost half of the instances of several counterparts.

As expected,
if we focus on the minimisation of erroneous instances and relations,
we identify GPT-based solutions
-- and especially GPT 3.5 --
as the best approach.
The large difference in the size of the GPT-based models against the selected counterparts represents a clear advantage
whenever the LLM is presented with the complex task of classifying instances into the corresponding semantic classes
or listing ingredients of recipes without incurring into hallucinations or slightly unreliable recipes.
Indeed,
the ontologies constructed relying on the GPT family also present the least relevant amount of worrying mistakes or hallucinations
(see \Cref{sssec:interesting_samples}).

However,
our results also show that relying on closed source models does not represent the only viable approach,
as few open source models achieve similarly acceptable level of performance.
The \emph{Llama 2} model constructs a rather small,
but quite effective ontology,
keeping both the number of incorrect
-- $E_{ii}$ --
and meaningless individuals
-- $E_{mi}$ --
low,
thus highlighting its potential.

Moreover,
relying on the MoE process,
the \emph{Nous Hermes} model allows a large ontology to be constructed
-- almost twice as big as the GPT 3.5 version --,
while having almost 90\% of the added individuals correct
-- surpassing GPT 4 Turbo --
and reaching $Q=85.1\%$.
Once again,
these results confirm the correlation between the dimensionality of the LLM model and the quality of the constructed ontology,
as \emph{Llama 2} and \emph{Nous Hermes} represent the largest open source models
(see \Cref{tab:llms_sizes}).

Finally,
one may notice quite a strong proportionality between the relative individual error
and the relative relation error measures.
Indeed,
LLMs achieving small $\mathit{RIE}$ tend to achieve a smaller $\mathit{RRE}$ as well.
So,
the ability of the underlying LLM to suggest instances of a specific class
or to classify correctly random instances as belonging to a specific class
looks like proportional to the LLM's ability of building correct relations between instances,
without proposing meaningless associations.
This finding represents a relevant discovery
as it suggests the generality of the LLMs capability of dealing with instances and relations between instances of an ontology.

\paragraph{QoS-Related Remarks}

\begin{table}
    \caption{
        QoS measurements (duration, number of queries, cost) of \llmfkg{} over different state-of-the-art LLMs.
        Each line describes the QoS of the experiment corresponding to the same line in \Cref{tab:performance}.
    }
    \centering
    \begin{tabular}{l | c c c} \toprule
        LLM & $\mathit{\Delta t}$ & $\mathit{N}$ & $\$$
        \\ \midrule
        GPT 3.5 Turbo \cite{gpt3-2020} & 15m 39s & 1236 & 0.06
        \\
        GPT 4 Turbo \cite{gpt4} & 58m 43s & 2414 & 2.16
        \\ \midrule
        Openchat \cite{wang2023openchat} & 11h 50m 25s & 4710 & n.a.
        \\
        Llama2 \cite{llama2-2023} & 5h 8m28s & 1990 & n.a.
        \\
        Mistral \cite{mistral} & 2h 19m 40s & 907 & n.a.
        \\
        Gemma \cite{gemini} & 4h 55m 2s & 1141 & n.a.
        \\ \midrule
        Mixtral \cite{mixtral} & 12h 51m 2s & 2620 & n.a.
        \\
        Nous Hermes & 14h 40m 3s & 6083 & n.a.
        \\ \bottomrule
    \end{tabular}
    \label{tab:qos}
\end{table}

As far as QoS is concerned,
we summarise the measurements of our experiments in \Cref{tab:qos}.
It is worth noticing how GPT-based models generally imply much quicker execution times ($\mathit{\Delta t}$)
-- mostly due to the looser rate limitations that OpenAI applies to paying users --,
at the price of a moderate expense in terms of USD dollars ($\$$).
Other models are usable via the Hugging Face API which can be freely used, 
at the price of tighter rate limitations,
which result in much longer runs for \llmfkg{}.

Another interesting remark is about the overall amount of queries ($\mathit{N}$),
which vary greatly among runs,
depending on the model of the LLM oracle.
As the models were subject to the same length limitations,
we speculate that such variability is due to the different verbosity of the LLMs' responses,
which essentially depends on the model's training data and architecture.

\subsubsection{Mistakes and Hallucinations Examples}\label{sssec:interesting_samples}
We here 
present and analyse few relevant examples of mistakes and hallucinations we found throughout our investigation.
We rank these examples from least to most relevant,
helping to shed some light on the characteristics of \llmfkg{}'s shortcomings.

\paragraph{Minor mistakes}
Most recipes of most models
-- especially smaller open source models --
have troubles with the identification of the proper spices that compose a recipe.
These types of mistakes can not be considered as the product of the LLM hallucinating,
as they do not resemble the LLM leaving the context of the recipe,
nor can really be considered as blatantly wrong information,
as the concept of a recipe varies a lot depending on personal preferences.
Therefore,
we consider similar mistakes as relevant,
yet minor mistakes,
which we consider as quickly identifiable after the ontology filling process is complete.
Similar mistakes include the identification of sauces as toppings and vice-versa,
or the misclassification of an instance which is fuzzy by itself%
---e.g., is peanut butter butter?

\paragraph{Mistakes due to the ontology structure}

Depending on the context and the capability of human designers,
the skeleton composing the initial empty ontology may be more or less complete.
This may cause issues whenever \llmfkg{} deals with fuzzy instances
that are not easy to classify directly into one of the available classes.
This may be the case of uncommon LLM suggestions or peculiar items.
For example,
`crocodile' and `alligator' are not the most common type of consumed meat,
however it is true that there exists few recipes based on these meat cuts.
Therefore,
whenever the LLM oracle suggests adding the `crocodile' and `alligator' instances to the ontology,
it is not perfectly clear where they should be added.
This results in the case of `crocodile' and `alligator' to be classified as instances
of the class `poultry-derived food' by the \emph{Nous Hermes} model.
While the incidence frequency can be decreased via a rigorous and complete definition of the starting empty ontology classes,
the fuzziness of the LLM suggestions should be better taken into account when dealing with ontology population
as the boundary between reasonable and unreasonable suggestion is tight.

\paragraph{LLM's hallucinations}

LLMs are known to suffer hallucinations
which may affect the suggestion of instances and relations to be added to the ontology under construction.
For example,
in our experiments we found the \emph{chemotherapy infusion} keyword
to be suggested by the \emph{Mistral} model
as an instance for the infusion class.
This represents an hallucination of the underlying LLM,
which when prompted for a list of infusions in the food domain
-- and also having the dietician imposition from its context --,
produced the chemotherapy infusion shot as a possible suggestion.
The automatic post-processing procedure in \llmfkg{} is unable to identify the proposal as a wrongful instance,
as it respects grammar principles
and it is also semantically valid,
therefore adding it to the ontology under construction.
Similarly,
\emph{pizza} is confidently suggested as an American recipe by the \emph{Mixtral} model,
although it originally from Italy.
Dealing with such issues is complex,
as it is not really possible to check reliably if all the suggestions of the queried LLMs are to be considered as valid or as hallucinations.
Therefore,
the complete removal of these issues is almost impossible.
One may think to query multiple times the LLM to check if the predicted suggestion persists or not
-- relying on the hypothesis that hallucinations are rare and that the model is not confidently predicting an hallucinated suggestion --
and add to the ontology only the confirmed predictions.
However,
this would require querying multiple times the LLM only to check the validity of each suggestion,
thus rendering the overall ontology filling process (more) inefficient.

\paragraph{Worrying hallucinations}

Most commonly,
hallucinations cause the output to be not coherent with the given context,
thus creating instances that should not belong to the ontology to be constructed.
However,
there exists a possibility for LLMs to provide factually false information
-- under the form of hallucinations --,
which however,
belongs to the given context.
This probably represents the worst possible scenario for the ontology construction process,
since it results in wrong assertions which are very complex to detect and may sometimes be dangerous.
For example,
in our experiments we found that,
in the `edible mushrooms' class,
\emph{Gemma}'s first suggestion is the \emph{Amanita muscaria}
which is 
poisonous. 
Being a mushroom,
the `Amanita muscaria' does not represent an out-of-context suggestion for the LLM,
but it still represents a completely false information to consider it as an \emph{edible} mushroom,
as its poisonousness should be known to the LLM.
We also confirm that the LLM was trained to recognise the Amanita muscaria as a poisonous mushroom:
we separately asked for information about this fact and
\emph{Gemma} confirmed that the Amanita muscaria is in fact poisonous.

\subsection{Comparison with related works}
\label{ssec:comparison}

Here we compare \llmfkg{} with the other methods for LLM-augmented knowledge graph construction
surveyed in \Cref{ssec:llm-kg}.
The comparison involves two major phases.
In the first one we compare methods on a per-feature basis,
and we later filter out the methods that operate too differently
for a performance-based comparison with \llmfkg{} to be meaningful.
We then compare the performance of \llmfkg{} with the remaining methods.

\subsubsection{Feature-based comparison}
\label{sssec:comparison-features}

Here we identify five desirable features to be achieved for an \emph{ideal} ontology population method.
We then compare state-of-the-art methods, as well as \llmfkg{}, against these features.

An ideal ontology population method should
\begin{featurelist}
    \item\label{item:f1} not require textual information as input from which to extract instances or relation
    -- but it should rather rely on the knowledge acquired by the LLM oracle during training --,
    hence follow a \emph{document-free} approach.
    It should also
    \item\label{item:f2} not require additional training or fine-tuning of the LLM oracle,
    hence follow a \emph{training-free} approach.
    Finally, it should
    \item\label{item:f3} support the \emph{construction} of the entire ontology,
    not just focussing on generating entities or relations (but not both);
    \item\label{item:f4} allow for user-provided \emph{prompt templates}
    to tailor LLM queries on the process on the domain at hand;
    and
    \item\label{item:f5} guarantee that the generated ontology remains \emph{consistent},
    meaning that its structural integrity is preserved by the population process.
\end{featurelist}
\Cref{tab:compare-features} presents the results of our feature-based comparison,
showing which features are satisfied by each approach surveyed in \Cref{ssec:llm-kg}.

The majority of surveyed approaches either rely on user-provided textual data to ease the mining process,
or require expensive model training to effectively construct KGs.
Moreover,
no surveyed approach guarantees the integrity of the generated ontology
-- as they target KGs, so they do not keep ontological constraints into account --,
leading to potentially inconsistent outcomes.
Therefore,
\llmfkg{} represents the only approach in which all requirements are satisfied,
providing a document- and training-free ontology population approach,
based on prompt templating,
which uses the LLM as an oracle and guarantees the consistency of the generated ontology,
while Harvest~\cite{HaoTTNSZXH23} is the second-best approach in terms of features satisfaction.

Accordingly,
Harvest turns out to be the only surveyed approach for which the comparison for \llmfkg{} seems fair.
Therefore,
in \Cref{sssec:comparison-performance} we compare the empirical performance of \llmfkg{} against Harvest.

\definecolor{darkgreen}{rgb}{0.0, 0.5, 0.0}
\newcommand{\ok}{{\color{darkgreen}\cmark}}
\newcommand{\ko}{{\color{red}\xmark}}

\begin{table}
    \caption{
        Feature-based comparison between \llmfkg{} and state-of-the-art KG construction methods.
        The arrow ($\rightarrow$) denotes the best-featured method from the literature,
        namely Harvest,
        which we compare with \llmfkg{} under a performance-based perspective.
    }
    \resizebox{\columnwidth}{!}{
    \centering
    \begin{tabular}{c | c c c c c} \toprule
        \textbf{Method} & \makecell{\textbf{Document}\\\textbf{Free \ref{item:f1}}} & \makecell{\textbf{Training}\\\textbf{Free \ref{item:f2}}} & \makecell{\textbf{Construction}\\\textbf{\ref{item:f3}}} & \makecell{\textbf{Prompt}\\\textbf{Templating \ref{item:f4}}} & \makecell{\textbf{Consistency}\\\textbf{\ref{item:f5}}} \\
        \midrule
        \cite{LuanEmnlp2018} & \ko & \ko & \ok & \ko & \ko \\
        \cite{comet-2019} & \ko & \ko & \ok & \ko & \ko \\
        \cite{end-to-end-kg-kumar-2020} & \ko & \ko & \ok & \ko & \ko \\
        \midrule
        \cite{llm-as-encoders-choi-2021} & \ko & \ko & \ko & \ko & \ko \\
        \cite{llm-as-encoders-wang-2021} & \ko & \ko & \ko & \ko & \ko \\
        \cite{entity-discovery-decao-2021} & \ko & \ko & \ok & \ko & \ko \\
        \cite{end-tp-end-kg-melnyk-2021} & \ko & \ko & \ok & \ko & \ko \\
        \midrule
        \cite{kg-complenion-saxena-2022} & \ko & \ko & \ko & \ko & \ko \\
        \cite{kg-complenion-chen-2022} & \ko & \ko & \ko & \ko & \ko \\
        \cite{kg-complenion-xin-2022} & \ko & \ko & \ko & \ko & \ko \\
        \cite{entity-discovery-ayoola-2022} & \ko & \ok & \ok & \ko & \ko \\
        \midrule
        \cite{llm-as-encoders-shen-2023} & \ko & \ko & \ko & \ko & \ko \\
        \cite{end-to-end-kg-han-2023} & \ko & \ok & \ok & \ok & \ko \\
        $\rightarrow$\cite{HaoTTNSZXH23} & \ok & \ok & \ok & \ok & \ko \\
        \midrule
        \cite{LieEacl2024} & \ok & \ko & \ok & \ok & \ko \\
        \cite{FanIpm2024} & \ok & \ko & \ok & \ok & \ko \\
        \midrule
        \midrule
        \makecell{\textbf{Ours}\\(\llmfkg)} & \ok & \ok & \ok & \ok & \ok \\
        \bottomrule
    \end{tabular}
    }
    \label{tab:compare-features}
\end{table}

\subsubsection{Performance-based comparison}
\label{sssec:comparison-performance}

\begin{table*}
    \caption{Performance of Harvest~\cite{HaoTTNSZXH23} over different state-of-the-art pre-trained models.}
    \resizebox{\textwidth}{!}{
    \centering
    \begin{tabular}{l | c c c c c c c c c c c c c c | c} \toprule
        LLM                                 & $\mathit{TI}$        & $\mathit{minCW}$     & $\mathit{maxCW}$    & $\mathit{TL}$        & $\mathit{TE}$          & $\mathit{RIE}$         & $E_{mis}$       & $E_{ii}$      & $E_{mi}$      & $E_{ci}$      & $E_{di}$      & $\mathit{TR}$       & $E_{wr}$       & $\mathit{RRE}$      & $Q$
        \\ \midrule
        RoBERTa base                        & 812	               & 2                    & 36                  & 682                  & 617                    & 0.7599                 & 391             & 118           & 84            & 70            & 20            & 40                  & 25             & 0.6250              & 0.2465
        \\
        RoBERTa large                       & 580                  & 0                    & 36                  & 867                  & 440                    & 0.7586                 & 479             & 126           & 54            & 30            & 25            & 55                  & 27             & 0.4909              & 0.2646
        \\
        BERT large cased                    & 1086                 & 22                   & 61                  & 942                  & 909                    & 0.8370                 & 734             & 153           & 45            & 97            & 125           & 51                  & 37             & 0.7255              & 0.1680
        \\ \bottomrule
    \end{tabular}
    }
    \label{tab:compare-performance}
\end{table*}

As highlighted in \Cref{sssec:comparison-features},
most state-of-the-art methods for LLM-augmented knowledge graph construction are indeed not directly comparable with \llmfkg{},
as they address the problem in substantially different ways;
and this is the reason why we focus on Harvest~\cite{HaoTTNSZXH23}.

Our performance-based comparison is straightforward.
We run Harvest to populate the same ontology schema presented in \Cref{ssec:ontology},
and we evaluate the populated ontology using the same performance metrics defined in \Cref{ssec:performance-metrics}.
In other words,
we evaluate Harvest's performance exactly as we did for \llmfkg{} in \Cref{ssec:results},
and we discuss the differences.
In doing so,
we replicate the experiments on several LLM families,
namely the ones where Harvest has been tested in the original paper~\cite{HaoTTNSZXH23}:
\begin{description}
    \item[RoBERTa base:] 12 layer, 768-hidden, 12-heads, 125M parameters RoBERTa~\cite{roberta-2019} using the BERT-base architecture;
    \item[RoBERTa large:] same architecture as above, but with 24 layers, 1024 hidden, 16 heads, and 355M parameters;
    \item[BERT large cased:] 24-layer, 1024-hidden, 16-heads, 340M parameters, BERT~\cite{bert-2019} trained on lower-cased English text.
\end{description}
The number followed by \emph{heads} and the number followed by \emph{hidden} refer to the number of attention heads and the hidden dimension of each block of the transformer architecture respectively.
Descriptions are taken from Hugging Face's documentation\footnotemark{}.
\footnotetext{\url{https://huggingface.co/transformers/v3.3.1/pretrained_models.html}}

\begin{figure}
    \centering
    \includegraphics[width=\linewidth]{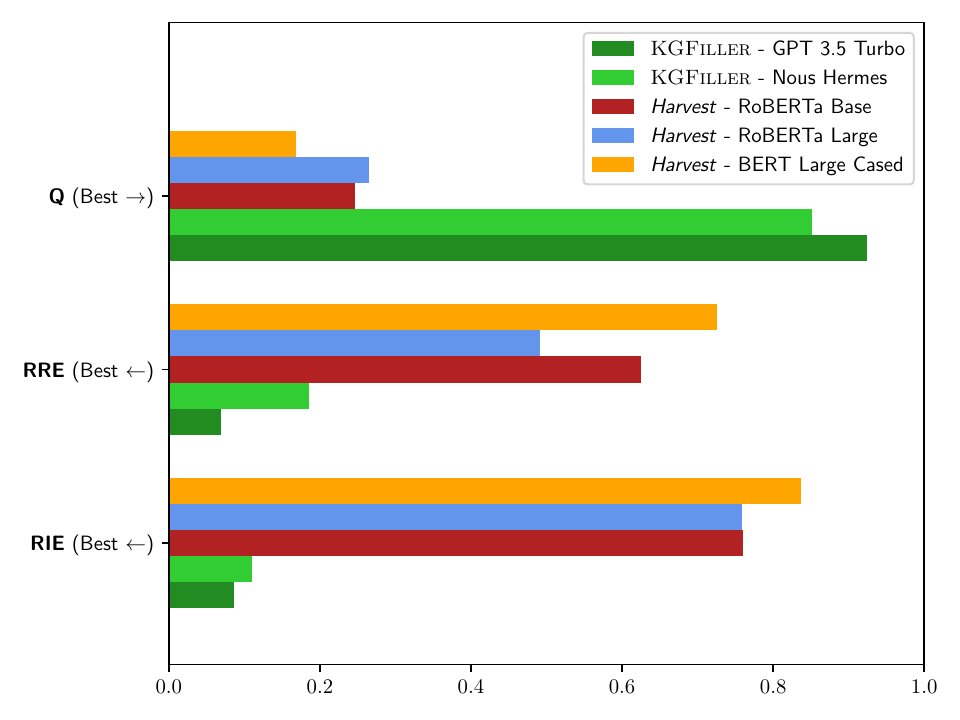}
    \caption{
        Comparison of the performance of Harvest~\cite{HaoTTNSZXH23} and \llmfkg{} w.r.t. the task of populating our food ontology (cf. \Cref{ssec:ontology}).
        For \llmfkg{}, the best-performing closed-source and open-source LLM models are considered---respectively GPT 3.5 Turbo and Nous Hermes.
    }
    \label{fig:compare-performance}
\end{figure}

The results of our experiments with Harvest are reported in \Cref{tab:compare-performance},
while the source code of the comparison is available on GitHub\footnotemark{}.
\footnotetext{\url{https://github.com/Chistera4-Expectation/experiments-knowledge-harvest-from-llm}}
Furthermore,
to ease the comparison between \llmfkg{} and Harvest,
we summarise the best-performing run of each method in \Cref{fig:compare-performance}.
As the reader can notice by comparing \Cref{tab:performance} and \Cref{tab:compare-performance},
or by looking at \Cref{fig:compare-performance},
\llmfkg{} outperforms Harvest in all the considered metrics.
The quality metric of the generated ontology is up to five times higher when using \llmfkg{} over Harvest, while the $RIE$ and $RRE$ are up to ten times lower.

\section{Discussion and SWOT Analysis}
\label{sec:discussion}

To the best of our knowledge,
our approach is an innovation in the field of ontology population,
and it impacts the way structured knowledge can generated to match a specific domain
and/or a schema provided by the user.
%

%
Here,
we propose a SWOT analysis
-- namely, a discussion about the \underline{s}trengths, \underline{w}eaknesses, \underline{o}pportunities, and \underline{t}hreats --
of \llmfkg{}.
Along this line,
we let our focus shift from the technical aspects of the approach
-- namely, \emph{how} (cf.\ \Cref{sec:contribution}) and \emph{whether} (cf.\ \Cref{sec:case-study}) it works --
to strategical aspects,
namely, why and when it should be used,
what are the possible future developments of the approach,
what are its current limitations,
and how they could be addressed in the future.

\subsection{Strengths}

\paragraph{Automation}

By relying on LLM oracles
-- which are trained on large amounts of general-knowledge data \emph{once and for all} --,
\llmfkg{} overcomes the limitations of state-of-the-art approaches for ontology population,
which require large \emph{corpora} of textual documents to be available,
from which to extract concepts, instances, and properties.
Therefore,
\llmfkg{} represents the first truly automatic ontology population procedure,
which does not require the user to provide any prior form of domain knowledge representation
to populate any given ontology.

\paragraph{Generality}

LLMs make \llmfkg{} a general-purpose mechanism to be applied to virtually any domain,
independently of data/domain knowledge availability.
Arguably,
\llmfkg{} is poorly affected by the narrowness of the domain,
as LLM oracles are usually trained on large amount of data gathered from the Web,
which cover most of the humankind's written knowledge.
On the other hand,
state-of-the-art ontology population approaches heavily suffer the domain-sensitivity issue,
as the corpus of documents they rely upon may be incomplete, non-representative, or simply lacking.
%


\paragraph{Incrementality}

\llmfkg{} does not require the input ontology to be empty.
In other words,
\llmfkg{} supports not only the population of the ontology,
but its refinement as well.
In this sense,
we say that \llmfkg{} is an \emph{incremental} approach,
which can be applied to already-instantiated ontologies,
aiming at enriching or populating them even more.
Along this line,
one may simply think to execute \llmfkg{} multiple times,
starting from an empty ontology and populating it incrementally.
%

\paragraph{Open future}

%
As LLMs have been demonstrated capable of going beyond the data they are trained on,
and they are expected to spontaneously develop more emergent capabilities as they grow in size and complexity,
we expect that the quality of the ontology population process will increase as well,
by simply relying on the latest LLM models available in the future.
%

\subsection{Weaknesses}

\paragraph{Completeness}

Similarly to state-of-the-art approaches,
\llmfkg{} does not provide completeness guarantees for the populated ontologies:
results may be lacking instances or relations that the user would expect to be present.
Arguably, this is intrinsic to any (semi-)automatic data-driven ontology population approach.
Put simply,
incompleteness is due to the fact that \llmfkg{} is not designed
to extract \emph{all} the knowledge the LLM oracle holds.
It rather provides means to limit the amount of instances extracted (to save time or money).
And,
even if it was designed to extract all the knowledge,
there is no guarantee that the LLM oracle has acquired all
mankind's knowledge in the first place.
%

\paragraph{Sampling bias}

Being completely dependent on what LLM oracle has learnt from the Web,
\llmfkg{} may be subject to sampling bias issues.
In fact,
the algorithm can only extract what the LLM oracle has learnt
(or what is inferable from that knowledge).
But, if the LLM did not learn about a specific topic,
and that topic is not inferable from what the LLM knows,
then \llmfkg{} will not be able to extract that topic.
As LLMs are trained on the Web,
under-represented topics on the Web may be less likely to be extracted by \llmfkg{},
thus leading to a possibly unbalanced or biased ontology.


\paragraph{Correctness}

Similarly to any (semi-)automatic data-driven ontology population approach,
\llmfkg{} does not provide any guarantee that the extracted information is correct.
In this context,
`correct' is intended as the truthfulness of the information added to the ontology.
In fact,
as exemplified in \Cref{sssec:interesting_samples},
LLM may mistakenly assign poisonous mushrooms to some class of edible mushrooms,
which is of course incorrect.
This is due to the fact that LLMs are prone to hallucinations and to providing false information%
---a well-known issue in the LLM community~\cite{hallucination-2023}.
%


\subsection{Opportunities}

\paragraph{Cold start}

The development workflow of any \emph{novel} data-driven or knowledge-based software system shall eventually
face the cold start issue:
the system is well-engineered and operative,
but for it to be effective there must be some data to operate upon.
But to gather data, the system should have users,
and to have users, the system should be effective.

%
In such situations,
relying on \llmfkg{}'s automatic ontology population procedure can help mitigating the cold start issue,
by providing a quick, cheap, and reliable way to
generate domain-specific data for the system to start being effective.

For instance,
in the \expectation{} project~\cite{expectation-extraamas2021},
a nutritional recommender system was needed (cf.~\cite{skerecommender-cmbp235}),
but not enough data about food ingredients and recipes was available:
that fuelled the idea of \llmfkg{}.

\paragraph{Automatic Hierarchy Construction}

\llmfkg{} focuses on ontology population.
As such,
it assumes the existence of a pre-defined ontology schema.
However,
extensions of the algorithm where LLM oracles are queried to construct the ontology schema itself are possible.
Approaches of this sort have been proposed in the literature,
e.g.\ \cite{funk2023ontology},
where the task is referred to as \emph{hierarchy construction}.

Along this line,
we envision further extensions of \llmfkg{},
aimed at automatising the whole ontology construction process,
e.g. by intertwining hierarchy construction and ontology population activities.
For instance,
the algorithm may be extended to identify lacking concepts in the input ontology
in order to better assign the individuals generated by the LLM oracle.
The result would be an intriguing continuous learning setup,
where domain knowledge is repetitively extended and refined.

\paragraph{Zero-Shot Ontology Creation}

Generalising upon the idea of automatic hierarchy construction,
one may wonder if \llmfkg{} can be extended to support the creation of ontologies from scratch.
%
Once a mechanism to construct the ontology schema is in place,
it would be interesting to explore if,
and to what extent,
\llmfkg{} can be used to populate an empty ontology.


\paragraph{Procedural Knowledge Extraction}

Another intriguing opportunity may arise from
the application of \llmfkg{} to the extraction of \emph{procedural} knowledge.
By procedural knowledge,
we refer to the structured data representing the steps required to complete a process,
or attain a specific goal.
In the fields of automated planning,
as well as multi-agent systems,
procedural knowledge is the way by which agents' \emph{plans} are encoded.
One key aspects of plans is that agents can \emph{execute} them,
to reproduce the process or to attain the goal.

What if \llmfkg{} is exploited to extract plans from LLM oracles?
This would pave the way towards the possibility,
for a software agent,
to learn novel behaviours autonomously,
by simply asking an LLM \emph{how} to do something,
and letting \llmfkg{} translate that in structured
-- hence executable --
form.

\subsection{Threats}

\paragraph{LLM performance degradation}

Web-based, conversational agents backed by LLM technology
-- especially OpenAI's ChatGPT --
have been shown to suffer from performance \emph{instability} over time \cite{ChenGptAndTime}.
This phenomenon is probably linked with the continuous optimisation process these agents periodically undergo.
Being trained not only on the Web,
but on previous interactions too,
the quality of their training data varies over time,
and the same is true for their performance.

Instability potentially affects \llmfkg{}'s performances as well.
Depending on the moment when the algorithm is executed,
the quality of the generated responses may vary significantly,
possibly hindering the reliability of the final ontology.


\paragraph{Lock-in effect}

%
As for any other technology provided `as a Service'
by the very few private players who can afford the infrastructure to train and maintain LLMs,
there is a chance for the lock-in effect~\cite{Eurich2014TheBL} to arise.
Once an ontology is populated using a specific LLM,
it may be difficult to consider leveraging a different LLM to refine or modify the created ontology,
as the knowledge of the second LLM would be (slightly?) different. 
Therefore,
users might end up in a situation where there is an over-reliance on a specific LLM technology,
a no other different technology could be exploited unless the whole ontology is re-populated from scratch,
which might be very expensive.

Combined with performance instability,
the lock-in effect may also affect the applicability of \llmfkg{} as a whole.
This may happen in the periods of time when the LLM oracle is not performing well:
in that period locked-in users may have no choice but to wait for the LLM to be re-trained,
and hope for the best.

\section{Conclusions}
\label{sec:conclusions}

Ontologies provide a structured framework that defines concepts, relationships, and properties within a specific domain,
defining an unambiguous and understandable representation of domain-specific knowledge.
However,
ontologies come along with the cost of a meticulous and time-consuming population process,
which is usually manual and possibly affected by humans’ errors and biases.
In this paper,
we hypothesise that LLMs incorporate a relevant amount of domain-specific knowledge obtained from their domain-agnostic training process,
involving huge amount of data gathered from the Web.
Accordingly,
we propose \llmfkg{},
a novel automatic ontology population approach leveraging LLMs.
Starting from
\begin{inlinelist}
    \item an initial schema composed by inter-related classes and properties,
    and on
    \item a set of query templates,
\end{inlinelist}
\llmfkg{} queries the LLM multiple times,
and generates instances for both classes, relationships, and properties from its replies.
Several other queries are performed to refine the ontology,
balancing the instances in the class hierarchy and avoiding the occurrence of duplicates.

\llmfkg{}'s validity is assessed through a case study on the food domain,
where the ontology population requires gathering food ingredients
-- over multiple categories --
and recipes.
We test \llmfkg{} over multiple state-of-the-art LLMs
-- open source, closed source and MoEs --,
measuring an array of metrics that analyse the quality of the constructed ontology.
Our results highlight \llmfkg{}'s ability of reliably populating a given ontology across several LLMs,
reaching instances and relations accuracy of up to $0.91$ and $0.93$ for the GPT 3.5 model,
respectively.
As expected,
larger LLMs achieve better performance,
while we highlight an interesting correlation between erroneous instances and misplaced relations.
Finally,
we analyse the various degrees of error severity,
showcasing how LLMs hallucinations may negatively impact the automatic ontology population process.
Overall,
the obtained results are encouraging,
as they show how the complex process of constructing ontologies can be
-- at least partially --
automated,
while still requiring few caveats and minor post-processing efforts.
%



\subsubsection*{CRediT Authors Statement}

Giovanni Ciatto: Conceptualisation, Methodology, Software, Validation, Investigation, Resources, Data Curation, Writing -- Original Draft, Writing -- Review and Editing, Supervision, Project Administration.
Andrea Agiollo: Methodology, Software, Validation, Investigation, Writing -- Original Draft, Writing -- Review and Editing, Visualisation.
Matteo Magnini: Conceptualisation, Methodology, Software, Investigation, Writing -- Original Draft, Writing -- Review and Editing.
Andrea Omicini: Writing -- Review and Editing, Funding Acquisition, Supervision, Project Administration.

\subsubsection*{Fundings}

This paper was partially supported by:
\begin{inlinelist}
   \item the CHIST-ERA IV project \expectation{} (\texttt{CHISTERA-19-XAI-005}), funded by the Italian MUR;

    \item the ``FAIR--Future Artificial Intelligence Research'' project
    (PNRR, M4C2, Investimento 1.3, Partenariato Esteso \texttt{PE00000013}),
    funded by the EC under the NextGenerationEU programme;

    \item the ``ENGINES--ENGineering INtElligent Systems around intelligent agent technologies'' project
    funded by the Italian MUR program ``PRIN 2022'' under G.A.\ \texttt{20229ZXBZM}.
\end{inlinelist}

\subsubsection*{Statements of ethical approval}

This study has been performed involving no patient, no volunteer, and no animal.
No personal or sensitive data has been exploited.

\subsubsection*{Declaration of interests}

The authors declare that they have no known competing financial interests or personal relationships that could have appeared to influence the work reported in this paper.

\subsubsection*{Declaration of generative AI in scientific writing}

During the preparation of this work,
the authors used GitHub's Copilot in order to speed-up their writing.
After using this tool,
the authors reviewed and edited the content as needed and take full responsibility for the content of the publication.

\bibliographystyle{elsarticle-num}
\bibliography{llm4kg-2023}

\begin{thebibliography}{10}
\expandafter\ifx\csname url\endcsname\relax
  \def\url#1{\texttt{#1}}\fi
\expandafter\ifx\csname urlprefix\endcsname\relax\def\urlprefix{URL }\fi
\expandafter\ifx\csname href\endcsname\relax
  \def\href#1#2{#2} \def\path#1{#1}\fi

\bibitem{Grimm10}
S.~Grimm, Knowledge representation and ontologies, in: M.~M. Gaber (Ed.), Scientific Data Mining and Knowledge Discovery - Principles and Foundations, Springer, 2010, pp. 111--137.
\newblock \href {https://doi.org/10.1007/978-3-642-02788-8_6} {\path{doi:10.1007/978-3-642-02788-8_6}}.

\bibitem{skeislr-acmcs}
G.~Ciatto, F.~Sabbatini, A.~Agiollo, M.~Magnini, A.~Omicini, Symbolic knowledge extraction and injection with sub-symbolic predictors: A systematic literature review, ACM Computing Surveys 56~(6) (2024) 161:1--161:35.
\newblock \href {https://doi.org/10.1145/3645103} {\path{doi:10.1145/3645103}}.

\bibitem{petasis-2011}
G.~Petasis, V.~Karkaletsis, G.~Paliouras, A.~Krithara, E.~Zavitsanos, Ontology population and enrichment: State of the art, in: G.~Paliouras, C.~D. Spyropoulos, G.~Tsatsaronis (Eds.), Knowledge-Driven Multimedia Information Extraction and Ontology Evolution - Bridging the Semantic Gap, Vol. 6050 of Lecture Notes in Computer Science, Springer, 2011, pp. 134--166.
\newblock \href {https://doi.org/10.1007/978-3-642-20795-2_6} {\path{doi:10.1007/978-3-642-20795-2_6}}.

\bibitem{BaaderHS08}
F.~Baader, I.~Horrocks, U.~Sattler, Description logics, in: F.~van Harmelen, V.~Lifschitz, B.~W. Porter (Eds.), Handbook of Knowledge Representation, Vol.~3 of Foundations of Artificial Intelligence, Elsevier, 2008, pp. 135--179.
\newblock \href {https://doi.org/10.1016/S1574-6526(07)03003-9} {\path{doi:10.1016/S1574-6526(07)03003-9}}.

\bibitem{bernerslee2001semantic}
T.~Berners-Lee, J.~Hendler, O.~Lassila, \href{https://www.scientificamerican.com/article/the-semantic-web/}{The semantic web}, Scientific American 284~(5) (2001) 34--43.
\newline\urlprefix\url{https://www.scientificamerican.com/article/the-semantic-web/}

\bibitem{Musen15}
M.~A. Musen, The {Prot{\'e}g{\'e}} project: a look back and a look forward, {AI} Matters 1~(4) (2015) 4--12.
\newblock \href {https://doi.org/10.1145/2757001.2757003} {\path{doi:10.1145/2757001.2757003}}.

\bibitem{Lamy17}
J.~Lamy, Owlready: Ontology-oriented programming in {P}ython with automatic classification and high level constructs for biomedical ontologies, Artificial Intelligence in Medicine 80 (2017) 11--28.
\newblock \href {https://doi.org/10.1016/J.ARTMED.2017.07.002} {\path{doi:10.1016/J.ARTMED.2017.07.002}}.

\bibitem{lubani-2019}
M.~Lubani, S.~A.~M. Noah, R.~Mahmud, Ontology population: Approaches and design aspects, J. Inf. Sci. 45~(4) (2019).
\newblock \href {https://doi.org/10.1177/0165551518801819} {\path{doi:10.1177/0165551518801819}}.

\bibitem{cherifa-2021}
A.~C. Khadir, H.~Aliane, A.~Guessoum, Ontology learning: Grand tour and challenges, Comput. Sci. Rev. 39 (2021) 100339.
\newblock \href {https://doi.org/10.1016/j.cosrev.2020.100339} {\path{doi:10.1016/j.cosrev.2020.100339}}.

\bibitem{Alaswadi-2020}
F.~N. Al{-}Aswadi, C.~H. Yong, K.~H. Gan, Automatic ontology construction from text: a review from shallow to deep learning trend, Artificial Intelligence Review 53~(6) (2020) 3901--3928.
\newblock \href {https://doi.org/10.1007/s10462-019-09782-9} {\path{doi:10.1007/s10462-019-09782-9}}.

\bibitem{finkestein-1999}
M.~Finkelstein-Landau, E.~Morin, Extracting semantic relationships between terms: Supervised {\it vs.} unsupervised methods, in: International Workshop on Ontological Engineering on the Global Information Infrastructure, 1999, pp. 71--80.

\bibitem{morin-1999}
E.~Morin, Automatic acquisition of semantic relations between terms from technical corpora, in: Proceedings of the Fifth International Congress on Terminology and Knowledge Engineering (TKE'99), 1999.

\bibitem{harith-2003}
A.~Harith, S.~Kim, D.~E. Millard, M.~J. Weal, W.~Hall, P.~H. Lewis, N.~R. Shadbolt, Automatic ontology-based knowledge extraction and tailored biography generation from the web, IEEE Intelligent Systems 18~(1) (2003) 14--21.

\bibitem{tanev-2006}
H.~Tanev, B.~Magnini, \href{https://aclanthology.org/E06-1003/}{Weakly supervised approaches for ontology population}, in: D.~McCarthy, S.~Wintner (Eds.), {EACL} 2006, 11st Conference of the European Chapter of the Association for Computational Linguistics, Proceedings of the Conference, April 3-7, 2006, Trento, Italy, The Association for Computer Linguistics, 2006.
\newline\urlprefix\url{https://aclanthology.org/E06-1003/}

\bibitem{yoon-2007}
H.~Yoon, Y.~Han, S.~Park, S.~Park, Ontology population from unstructured and semi-structured texts, in: Proceedings of The Sixth International Conference on Advanced Language Processing and Web Information Technology, {ALPIT} 2007, Luoyang, Henan, China, 22-24 August 2007, {IEEE} Computer Society, 2007, pp. 135--139.
\newblock \href {https://doi.org/10.1109/ALPIT.2007.30} {\path{doi:10.1109/ALPIT.2007.30}}.

\bibitem{maynard-2008}
D.~Maynard, Y.~Li, W.~Peters, \href{https://ebooks.iospress.nl/volume/proof-technology-and-computation}{{NLP} techniques for term extraction and ontology population}, in: P.~Buitelaar, P.~Cimiano (Eds.), Ontology Learning and Population: Bridging the Gap between Text and Knowledge, Vol. 167 of Frontiers in Artificial Intelligence and Applications, {IOS} Press, 2008, pp. 107--127.
\newline\urlprefix\url{https://ebooks.iospress.nl/volume/proof-technology-and-computation}

\bibitem{celjuska-2004}
D.~Celjuska, M.~Vargas-Vera, Ontosophie: A semi-automatic system for ontology population from text, in: International Conference on Natural Language Processing (ICON), Vol.~60, 2004.

\bibitem{etzioni-2005}
O.~Etzioni, M.~J. Cafarella, D.~Downey, A.~Popescu, T.~Shaked, S.~Soderland, D.~S. Weld, A.~Yates, Unsupervised named-entity extraction from the web: An experimental study, Artificial Intelligence 165~(1) (2005) 91--134.
\newblock \href {https://doi.org/10.1016/j.artint.2005.03.001} {\path{doi:10.1016/j.artint.2005.03.001}}.

\bibitem{jiang-2011}
M.~Jiang, et~al., A study of machine-learning-based approaches to extract clinical entities and their assertions from discharge summaries, J. Am. Medical Informatics Assoc. 18~(5) (2011) 601--606.
\newblock \href {https://doi.org/10.1136/amiajnl-2011-000163} {\path{doi:10.1136/amiajnl-2011-000163}}.

\bibitem{souili-2015}
A.~Souili, D.~Cavallucci, F.~Rousselot, Natural language processing (nlp)--a solution for knowledge extraction from patent unstructured data, Procedia engineering 131 (2015) 635--643.

\bibitem{liu-2013}
C.~Liu, W.~Sun, W.~Chao, W.~Che, Convolution neural network for relation extraction, in: H.~Motoda, Z.~Wu, L.~Cao, O.~R. Za{\"{\i}}ane, M.~Yao, W.~Wang (Eds.), Advanced Data Mining and Applications - 9th International Conference, {ADMA} 2013, Hangzhou, China, December 14-16, 2013, Proceedings, Part {II}, Vol. 8347 of Lecture Notes in Computer Science, Springer, 2013, pp. 231--242.
\newblock \href {https://doi.org/10.1007/978-3-642-53917-6_21} {\path{doi:10.1007/978-3-642-53917-6_21}}.

\bibitem{zeng-2014}
D.~Zeng, K.~Liu, S.~Lai, G.~Zhou, J.~Zhao, \href{https://aclanthology.org/C14-1220/}{Relation classification via convolutional deep neural network}, in: J.~Hajic, J.~Tsujii (Eds.), {COLING} 2014, 25th International Conference on Computational Linguistics, Proceedings of the Conference: Technical Papers, August 23-29, 2014, Dublin, Ireland, {ACL}, 2014, pp. 2335--2344.
\newline\urlprefix\url{https://aclanthology.org/C14-1220/}

\bibitem{ayadi-2019}
A.~Ayadi, A.~Samet, F.~de~Bertrand~de Beuvron, C.~Zanni{-}Merk, Ontology population with deep learning-based {NLP:} a case study on the biomolecular network ontology, in: I.~J. Rudas, J.~Csirik, C.~Toro, J.~Botzheim, R.~J. Howlett, L.~C. Jain (Eds.), Knowledge-Based and Intelligent Information {\&} Engineering Systems: Proceedings of the 23rd International Conference KES-2019, Budapest, Hungary, 4-6 September 2019, Vol. 159 of Procedia Computer Science, Elsevier, 2019, pp. 572--581.
\newblock \href {https://doi.org/10.1016/j.procs.2019.09.212} {\path{doi:10.1016/j.procs.2019.09.212}}.

\bibitem{bert-2019}
J.~Devlin, M.~Chang, K.~Lee, K.~Toutanova, {BERT:} pre-training of deep bidirectional transformers for language understanding, in: J.~Burstein, C.~Doran, T.~Solorio (Eds.), Proceedings of the 2019 Conference of the North American Chapter of the Association for Computational Linguistics: Human Language Technologies, {NAACL-HLT} 2019, Minneapolis, MN, USA, June 2-7, 2019, Volume 1 (Long and Short Papers), Association for Computational Linguistics, 2019, pp. 4171--4186.
\newblock \href {https://doi.org/10.18653/v1/n19-1423} {\path{doi:10.18653/v1/n19-1423}}.

\bibitem{gpt3-2020}
T.~B. Brown, et~al., \href{https://proceedings.neurips.cc/paper/2020/hash/1457c0d6bfcb4967418bfb8ac142f64a-Abstract.html}{Language models are few-shot learners}, in: H.~Larochelle, M.~Ranzato, R.~Hadsell, M.~Balcan, H.~Lin (Eds.), Advances in Neural Information Processing Systems 33: Annual Conference on Neural Information Processing Systems 2020, NeurIPS 2020, December 6-12, 2020, virtual, 2020.
\newline\urlprefix\url{https://proceedings.neurips.cc/paper/2020/hash/1457c0d6bfcb4967418bfb8ac142f64a-Abstract.html}

\bibitem{MinRSSVNAHR23}
B.~Min, H.~Ross, E.~Sulem, A.~P.~B. Veyseh, T.~H. Nguyen, O.~Sainz, E.~Agirre, I.~Heintz, D.~Roth, Recent advances in natural language processing via large pre-trained language models: A survey, ACM Computing Surveys 56~(2) (sep 2023).
\newblock \href {https://doi.org/10.1145/3605943} {\path{doi:10.1145/3605943}}.

\bibitem{PetroniRRLBWM19}
F.~Petroni, T.~Rockt{\"{a}}schel, S.~Riedel, P.~S.~H. Lewis, A.~Bakhtin, Y.~Wu, A.~H. Miller, Language models as knowledge bases?, in: K.~Inui, J.~Jiang, V.~Ng, X.~Wan (Eds.), Proceedings of the 2019 Conference on Empirical Methods in Natural Language Processing and the 9th International Joint Conference on Natural Language Processing, {EMNLP-IJCNLP} 2019, Hong Kong, China, November 3-7, 2019, Association for Computational Linguistics, 2019, pp. 2463--2473.
\newblock \href {https://doi.org/10.18653/V1/D19-1250} {\path{doi:10.18653/V1/D19-1250}}.

\bibitem{ZhuKZSUTF15}
Y.~Zhu, R.~Kiros, R.~S. Zemel, R.~Salakhutdinov, R.~Urtasun, A.~Torralba, S.~Fidler, Aligning books and movies: Towards story-like visual explanations by watching movies and reading books, in: 2015 {IEEE} International Conference on Computer Vision, {ICCV} 2015, Santiago, Chile, December 7-13, 2015, {IEEE} Computer Society, 2015, pp. 19--27.
\newblock \href {https://doi.org/10.1109/ICCV.2015.11} {\path{doi:10.1109/ICCV.2015.11}}.

\bibitem{gpt2-2019}
A.~Radford, J.~Wu, R.~Child, D.~Luan, D.~Amodei, I.~Sutskever, Language models are unsupervised multitask learners, \url{https://d4mucfpksywv.cloudfront.net/better-language-models/language-models.pdf} (2019).

\bibitem{training-data-attack-2021}
N.~Carlini, et~al., \href{https://www.usenix.org/conference/usenixsecurity21/presentation/carlini-extracting}{Extracting training data from large language models}, in: M.~Bailey, R.~Greenstadt (Eds.), 30th {USENIX} Security Symposium, {USENIX} Security 2021, August 11-13, 2021, {USENIX} Association, 2021, pp. 2633--2650.
\newline\urlprefix\url{https://www.usenix.org/conference/usenixsecurity21/presentation/carlini-extracting}

\bibitem{roberta-2019}
Y.~Liu, M.~Ott, N.~Goyal, J.~Du, M.~Joshi, D.~Chen, O.~Levy, M.~Lewis, L.~Zettlemoyer, V.~Stoyanov, \href{http://arxiv.org/abs/1907.11692}{Roberta: {A} robustly optimized {BERT} pretraining approach}, CoRR abs/1907.11692 (2019).
\newline\urlprefix\url{http://arxiv.org/abs/1907.11692}

\bibitem{Gokaslan2019OpenWeb}
A.~Gokaslan, V.~Cohen, Openwebtext corpus, \url{http://Skylion007.github.io/OpenWebTextCorpus} (2019).

\bibitem{TrieuQuoc2018}
T.~H. Trinh, Q.~V. Le, \href{http://arxiv.org/abs/1806.02847}{A simple method for commonsense reasoning}, CoRR abs/1806.02847 (2018).
\newline\urlprefix\url{http://arxiv.org/abs/1806.02847}

\bibitem{llama1-2023}
H.~Touvron, et~al., Llama: Open and efficient foundation language models, CoRR abs/2302.13971 (2023).
\newblock \href {https://doi.org/10.48550/ARXIV.2302.13971} {\path{doi:10.48550/ARXIV.2302.13971}}.

\bibitem{llama2-2023}
H.~Touvron, et~al., Llama 2: Open foundation and fine-tuned chat models, CoRR abs/2307.09288 (2023).
\newblock \href {https://doi.org/10.48550/ARXIV.2307.09288} {\path{doi:10.48550/ARXIV.2307.09288}}.

\bibitem{RaffelSRLNMZLL20}
C.~Raffel, N.~Shazeer, A.~Roberts, K.~Lee, S.~Narang, M.~Matena, Y.~Zhou, W.~Li, P.~J. Liu, \href{http://jmlr.org/papers/v21/20-074.html}{Exploring the limits of transfer learning with a unified text-to-text transformer}, J. Mach. Learn. Res. 21 (2020) 140:1--140:67.
\newline\urlprefix\url{http://jmlr.org/papers/v21/20-074.html}

\bibitem{hallucination-2023}
Y.~Zhang, et~al., Siren's song in the {AI} ocean: {A} survey on hallucination in large language models, CoRR abs/2309.01219 (2023).
\newblock \href {http://arxiv.org/abs/2309.01219} {\path{arXiv:2309.01219}}.

\bibitem{MemmertCB24}
L.~Memmert, I.~Cvetkovic, E.~A.~C. Bittner, \href{https://hdl.handle.net/10125/107289}{The more is not the merrier: Effects of prompt engineering on the quality of ideas generated by {GPT-3}}, in: T.~X. Bui (Ed.), 57th Hawaii International Conference on System Sciences, {HICSS} 2024, Hilton Hawaiian Village Waikiki Beach Resort, Hawaii, USA, January 3-6, 2024, ScholarSpace, 2024, pp. 7520--7529.
\newline\urlprefix\url{https://hdl.handle.net/10125/107289}

\bibitem{WeiTBRZBYBZMCHVLDF22}
J.~Wei, et~al., \href{https://openreview.net/forum?id=yzkSU5zdwD}{Emergent abilities of large language models}, Trans. Mach. Learn. Res. 2022 (2022).
\newline\urlprefix\url{https://openreview.net/forum?id=yzkSU5zdwD}

\bibitem{SchaefferMK23}
R.~Schaeffer, B.~Miranda, S.~Koyejo, Are emergent abilities of large language models a mirage?, CoRR abs/2304.15004 (2023).
\newblock \href {https://doi.org/10.48550/ARXIV.2304.15004} {\path{doi:10.48550/ARXIV.2304.15004}}.

\bibitem{WuQRAKKA23}
Z.~Wu, L.~Qiu, A.~Ross, E.~Aky{\"{u}}rek, B.~Chen, B.~Wang, N.~Kim, J.~Andreas, Y.~Kim, Reasoning or reciting? exploring the capabilities and limitations of language models through counterfactual tasks, CoRR abs/2307.02477 (2023).
\newblock \href {https://doi.org/10.48550/ARXIV.2307.02477} {\path{doi:10.48550/ARXIV.2307.02477}}.

\bibitem{roadmap-kg-2024}
S.~Pan, L.~Luo, Y.~Wang, C.~Chen, J.~Wang, X.~Wu, Unifying large language models and knowledge graphs: {A} roadmap, {IEEE} Trans. Knowl. Data Eng. 36~(7) (2024) 3580--3599.
\newblock \href {https://doi.org/10.1109/TKDE.2024.3352100} {\path{doi:10.1109/TKDE.2024.3352100}}.

\bibitem{ZhuWCQOYDCZ24}
Y.~Zhu, X.~Wang, J.~Chen, S.~Qiao, Y.~Ou, Y.~Yao, S.~Deng, H.~Chen, N.~Zhang, {LLMs} for knowledge graph construction and reasoning: recent capabilities and future opportunities, World Wide Web {(WWW)} 27~(5) (2024) 58.
\newblock \href {https://doi.org/10.1007/S11280-024-01297-W} {\path{doi:10.1007/S11280-024-01297-W}}.

\bibitem{PanRKSCDJO0LBMB23}
J.~Z. Pan, S.~Razniewski, J.~Kalo, S.~Singhania, J.~Chen, S.~Dietze, H.~Jabeen, J.~Omeliyanenko, W.~Zhang, M.~Lissandrini, R.~Biswas, G.~de~Melo, A.~Bonifati, E.~Vakaj, M.~Dragoni, D.~Graux, Large language models and knowledge graphs: Opportunities and challenges, {TGDK} 1~(1) (2023) 2:1--2:38.
\newblock \href {https://doi.org/10.4230/TGDK.1.1.2} {\path{doi:10.4230/TGDK.1.1.2}}.

\bibitem{kg-vs-ontology-2016}
L.~Ehrlinger, W.~W{\"{o}}{\ss}, \href{https://ceur-ws.org/Vol-1695/paper4.pdf}{Towards a definition of knowledge graphs}, in: M.~Martin, M.~Cuquet, E.~Folmer (Eds.), Joint Proceedings of the Posters and Demos Track of the 12th International Conference on Semantic Systems - SEMANTiCS2016 and the 1st International Workshop on Semantic Change {\&} Evolving Semantics (SuCCESS'16) co-located with the 12th International Conference on Semantic Systems (SEMANTiCS 2016), Leipzig, Germany, September 12-15, 2016, Vol. 1695 of {CEUR} Workshop Proceedings, CEUR-WS.org, 2016.
\newline\urlprefix\url{https://ceur-ws.org/Vol-1695/paper4.pdf}

\bibitem{llm-as-encoders-choi-2021}
B.~Choi, D.~Jang, Y.~Ko, {MEM-KGC:} masked entity model for knowledge graph completion with pre-trained language model, {IEEE} Access 9 (2021) 132025--132032.
\newblock \href {https://doi.org/10.1109/ACCESS.2021.3113329} {\path{doi:10.1109/ACCESS.2021.3113329}}.

\bibitem{llm-as-encoders-wang-2021}
B.~Wang, T.~Shen, G.~Long, T.~Zhou, Y.~Wang, Y.~Chang, Structure-augmented text representation learning for efficient knowledge graph completion, in: J.~Leskovec, M.~Grobelnik, M.~Najork, J.~Tang, L.~Zia (Eds.), {WWW} '21: The Web Conference 2021, Virtual Event / Ljubljana, Slovenia, April 19-23, 2021, {ACM} / {IW3C2}, 2021, pp. 1737--1748.
\newblock \href {https://doi.org/10.1145/3442381.3450043} {\path{doi:10.1145/3442381.3450043}}.

\bibitem{llm-as-encoders-shen-2023}
J.~Shen, C.~Wang, L.~Gong, D.~Song, \href{https://aclanthology.org/2022.coling-1.171}{Joint language semantic and structure embedding for knowledge graph completion}, in: N.~Calzolari, C.~Huang, H.~Kim, J.~Pustejovsky, L.~Wanner, K.~Choi, P.~Ryu, H.~Chen, L.~Donatelli, H.~Ji, S.~Kurohashi, P.~Paggio, N.~Xue, S.~Kim, Y.~Hahm, Z.~He, T.~K. Lee, E.~Santus, F.~Bond, S.~Na (Eds.), Proceedings of the 29th International Conference on Computational Linguistics, {COLING} 2022, Gyeongju, Republic of Korea, October 12-17, 2022, International Committee on Computational Linguistics, 2022, pp. 1965--1978.
\newline\urlprefix\url{https://aclanthology.org/2022.coling-1.171}

\bibitem{kg-complenion-saxena-2022}
A.~Saxena, A.~Kochsiek, R.~Gemulla, Sequence-to-sequence knowledge graph completion and question answering, in: S.~Muresan, P.~Nakov, A.~Villavicencio (Eds.), Proceedings of the 60th Annual Meeting of the Association for Computational Linguistics (Volume 1: Long Papers), {ACL} 2022, Dublin, Ireland, May 22-27, 2022, Association for Computational Linguistics, 2022, pp. 2814--2828.
\newblock \href {https://doi.org/10.18653/V1/2022.ACL-LONG.201} {\path{doi:10.18653/V1/2022.ACL-LONG.201}}.

\bibitem{kg-complenion-chen-2022}
C.~Chen, Y.~Wang, B.~Li, K.~Lam, \href{https://aclanthology.org/2022.coling-1.352}{Knowledge is flat: {A} seq2seq generative framework for various knowledge graph completion}, in: N.~Calzolari, C.~Huang, H.~Kim, J.~Pustejovsky, L.~Wanner, K.~Choi, P.~Ryu, H.~Chen, L.~Donatelli, H.~Ji, S.~Kurohashi, P.~Paggio, N.~Xue, S.~Kim, Y.~Hahm, Z.~He, T.~K. Lee, E.~Santus, F.~Bond, S.~Na (Eds.), Proceedings of the 29th International Conference on Computational Linguistics, {COLING} 2022, Gyeongju, Republic of Korea, October 12-17, 2022, International Committee on Computational Linguistics, 2022, pp. 4005--4017.
\newline\urlprefix\url{https://aclanthology.org/2022.coling-1.352}

\bibitem{kg-complenion-xin-2022}
X.~Xie, N.~Zhang, Z.~Li, S.~Deng, H.~Chen, F.~Xiong, M.~Chen, H.~Chen, From discrimination to generation: Knowledge graph completion with generative transformer, in: F.~Laforest, R.~Troncy, E.~Simperl, D.~Agarwal, A.~Gionis, I.~Herman, L.~M{\'{e}}dini (Eds.), Companion of The Web Conference 2022, Virtual Event / Lyon, France, April 25 - 29, 2022, {ACM}, 2022, pp. 162--165.
\newblock \href {https://doi.org/10.1145/3487553.3524238} {\path{doi:10.1145/3487553.3524238}}.

\bibitem{kg-complenion-zhu-2023}
Y.~Zhu, X.~Wang, J.~Chen, S.~Qiao, Y.~Ou, Y.~Yao, S.~Deng, H.~Chen, N.~Zhang, {LLMs} for knowledge graph construction and reasoning: Recent capabilities and future opportunities, CoRR abs/2305.13168 (2023).
\newblock \href {http://arxiv.org/abs/2305.13168} {\path{arXiv:2305.13168}}.

\bibitem{entity-discovery-ayoola-2022}
T.~Ayoola, S.~Tyagi, J.~Fisher, C.~Christodoulopoulos, A.~Pierleoni, {ReFinED}: An efficient zero-shot-capable approach to end-to-end entity linking, in: A.~Loukina, R.~Gangadharaiah, B.~Min (Eds.), Proceedings of the 2022 Conference of the North American Chapter of the Association for Computational Linguistics: Human Language Technologies: Industry Track, {NAACL} 2022, Hybrid: Seattle, Washington, {USA} + Online, July 10-15, 2022, Association for Computational Linguistics, 2022, pp. 209--220.
\newblock \href {https://doi.org/10.18653/V1/2022.NAACL-INDUSTRY.24} {\path{doi:10.18653/V1/2022.NAACL-INDUSTRY.24}}.

\bibitem{entity-discovery-decao-2021}
N.~D. Cao, G.~Izacard, S.~Riedel, F.~Petroni, \href{https://openreview.net/forum?id=5k8F6UU39V}{Autoregressive entity retrieval}, in: 9th International Conference on Learning Representations, {ICLR} 2021, Virtual Event, Austria, May 3-7, 2021, OpenReview.net, 2021.
\newline\urlprefix\url{https://openreview.net/forum?id=5k8F6UU39V}

\bibitem{end-to-end-kg-kumar-2020}
A.~Kumar, A.~Pandey, R.~Gadia, M.~Mishra, Building knowledge graph using pre-trained language model for learning entity-aware relationships, in: 2020 IEEE International Conference on Computing, Power and Communication Technologies (GUCON), 2020, pp. 310--315.
\newblock \href {https://doi.org/10.1109/GUCON48875.2020.9231227} {\path{doi:10.1109/GUCON48875.2020.9231227}}.

\bibitem{end-tp-end-kg-melnyk-2021}
I.~Melnyk, P.~Dognin, P.~Das, Grapher: Multi-stage knowledge graph construction using pretrained language models, in: NeurIPS 2021 Workshop on Deep Generative Models and Downstream Applications, 2021.

\bibitem{end-to-end-kg-han-2023}
J.~Han, N.~Collier, W.~L. Buntine, E.~Shareghi, Pive: Prompting with iterative verification improving graph-based generative capability of llms, CoRR abs/2305.12392 (2023).
\newblock \href {http://arxiv.org/abs/2305.12392} {\path{arXiv:2305.12392}}.

\bibitem{comet-2019}
A.~Bosselut, H.~Rashkin, M.~Sap, C.~Malaviya, A.~Celikyilmaz, Y.~Choi, {COMET:} commonsense transformers for automatic knowledge graph construction, in: A.~Korhonen, D.~R. Traum, L.~M{\`{a}}rquez (Eds.), Proceedings of the 57th Conference of the Association for Computational Linguistics, {ACL} 2019, Florence, Italy, July 28- August 2, 2019, Volume 1: Long Papers, Association for Computational Linguistics, 2019, pp. 4762--4779.
\newblock \href {https://doi.org/10.18653/V1/P19-1470} {\path{doi:10.18653/V1/P19-1470}}.

\bibitem{HaoTTNSZXH23}
S.~Hao, B.~Tan, K.~Tang, B.~Ni, X.~Shao, H.~Zhang, E.~P. Xing, Z.~Hu, Bertnet: Harvesting knowledge graphs with arbitrary relations from pretrained language models, in: A.~Rogers, J.~L. Boyd{-}Graber, N.~Okazaki (Eds.), Findings of the Association for Computational Linguistics: {ACL} 2023, Toronto, Canada, July 9-14, 2023, Association for Computational Linguistics, 2023, pp. 5000--5015.
\newblock \href {https://doi.org/10.18653/V1/2023.FINDINGS-ACL.309} {\path{doi:10.18653/V1/2023.FINDINGS-ACL.309}}.

\bibitem{ZhaoTkde2022}
X.~Zhao, W.~Zeng, J.~Tang, W.~Wang, F.~M. Suchanek, An experimental study of state-of-the-art entity alignment approaches, {IEEE} Transactions on Knowledge and Data Engineering 34~(6) (2022) 2610--2625.
\newblock \href {https://doi.org/10.1109/TKDE.2020.3018741} {\path{doi:10.1109/TKDE.2020.3018741}}.

\bibitem{ChenIjcai2018}
M.~Chen, Y.~Tian, K.~Chang, S.~Skiena, C.~Zaniolo, Co-training embeddings of knowledge graphs and entity descriptions for cross-lingual entity alignment, in: Proceedings of the Twenty-Seventh International Joint Conference on Artificial Intelligence, {IJCAI} 2018, July 13-19, 2018, Stockholm, Sweden, ijcai.org, 2018, pp. 3998--4004.
\newblock \href {https://doi.org/10.24963/IJCAI.2018/556} {\path{doi:10.24963/IJCAI.2018/556}}.

\bibitem{XuAcl2019}
K.~Xu, L.~Wang, M.~Yu, Y.~Feng, Y.~Song, Z.~Wang, D.~Yu, Cross-lingual knowledge graph alignment via graph matching neural network, in: Proceedings of the 57th Conference of the Association for Computational Linguistics, {ACL} 2019, Florence, Italy, July 28- August 2, 2019, Volume 1: Long Papers, Association for Computational Linguistics, 2019, pp. 3156--3161.
\newblock \href {https://doi.org/10.18653/V1/P19-1304} {\path{doi:10.18653/V1/P19-1304}}.

\bibitem{expectation-extraamas2021}
D.~Calvaresi, G.~Ciatto, A.~Najjar, R.~Aydo{\u g}an, L.~Van~der Torre, A.~Omicini, M.~I. Schumacher, {{\sc Expectation}}: Personalized explainable artificial intelligence for decentralized agents with heterogeneous knowledge, in: D.~Calvaresi, A.~Najjar, M.~Winikoff, K.~Fr{\"a}mling (Eds.), Explainable and Transparent AI and Multi-Agent Systems. Third International Workshop, EXTRAAMAS 2021, Virtual Event, May 3--7, 2021, Revised Selected Papers, Vol. 12688 of Lecture Notes in Computer Science, Springer Nature, Basel, Switzerland, 2021, pp. 331--343.
\newblock \href {https://doi.org/10.1007/978-3-030-82017-6_20} {\path{doi:10.1007/978-3-030-82017-6_20}}.

\bibitem{skerecommender-cmbp235}
M.~Magnini, G.~Ciatto, F.~Cant{\"u}rk, R.~Aydo{\v g}an, A.~Omicini, Symbolic knowledge extraction for explainable nutritional recommenders, Computer Methods and Programs in Biomedicine 235 (Jun. 2023).
\newblock \href {https://doi.org/10.1016/j.cmpb.2023.107536} {\path{doi:10.1016/j.cmpb.2023.107536}}.

\bibitem{wang2023openchat}
G.~Wang, S.~Cheng, X.~Zhan, X.~Li, S.~Song, Y.~Liu, Openchat: Advancing open-source language models with mixed-quality data, arXiv preprint arXiv:2309.11235 (2023).

\bibitem{mistral}
A.~Q. Jiang, et~al., Mistral 7b, CoRR abs/2310.06825 (2023).
\newblock \href {https://doi.org/10.48550/ARXIV.2310.06825} {\path{doi:10.48550/ARXIV.2310.06825}}.

\bibitem{gemini}
R.~Anil, et~al., Gemini: {A} family of highly capable multimodal models, CoRR abs/2312.11805 (2023).
\newblock \href {https://doi.org/10.48550/ARXIV.2312.11805} {\path{doi:10.48550/ARXIV.2312.11805}}.

\bibitem{mixtral}
A.~Q. Jiang, et~al., Mixtral of experts, CoRR abs/2401.04088 (2024).
\newblock \href {https://doi.org/10.48550/ARXIV.2401.04088} {\path{doi:10.48550/ARXIV.2401.04088}}.

\bibitem{gpt4}
OpenAI, {GPT-4} technical report, CoRR abs/2303.08774 (2023).
\newblock \href {https://doi.org/10.48550/ARXIV.2303.08774} {\path{doi:10.48550/ARXIV.2303.08774}}.

\bibitem{LuanEmnlp2018}
Y.~Luan, L.~He, M.~Ostendorf, H.~Hajishirzi, Multi-task identification of entities, relations, and coreference for scientific knowledge graph construction, in: Proceedings of the 2018 Conference on Empirical Methods in Natural Language Processing, Brussels, Belgium, October 31 - November 4, 2018, Association for Computational Linguistics, 2018, pp. 3219--3232.
\newblock \href {https://doi.org/10.18653/V1/D18-1360} {\path{doi:10.18653/V1/D18-1360}}.

\bibitem{LieEacl2024}
D.~Li, Z.~Tan, T.~Chen, H.~Liu, \href{https://aclanthology.org/2024.findings-eacl.32}{Contextualization distillation from large language model for knowledge graph completion}, in: Findings of the Association for Computational Linguistics: {EACL} 2024, St. Julian's, Malta, March 17-22, 2024, Association for Computational Linguistics, 2024, pp. 458--477.
\newline\urlprefix\url{https://aclanthology.org/2024.findings-eacl.32}

\bibitem{FanIpm2024}
Z.~Fan, C.~Chen, Cupe-kg: Cultural perspective-based knowledge graph construction of tourism resources via pretrained language models, Inf. Process. Manag. 61~(2) (2024) 103646.
\newblock \href {https://doi.org/10.1016/J.IPM.2024.103646} {\path{doi:10.1016/J.IPM.2024.103646}}.

\bibitem{funk2023ontology}
M.~Funk, S.~Hosemann, J.~C. Jung, C.~Lutz, Towards ontology construction with language models (2023).
\newblock \href {http://arxiv.org/abs/2309.09898} {\path{arXiv:2309.09898}}.

\bibitem{ChenGptAndTime}
L.~Chen, M.~Zaharia, J.~Zou, How is chatgpt's behavior changing over time?, CoRR abs/2307.09009 (2023).
\newblock \href {https://doi.org/10.48550/ARXIV.2307.09009} {\path{doi:10.48550/ARXIV.2307.09009}}.

\bibitem{Eurich2014TheBL}
M.~Eurich, M.~J. Burtscher, \href{https://cambridgeservicealliance.eng.cam.ac.uk/system/files/documents/2014AugustPaperBusinesstoConsumerLockinEffect.pdf}{The business-to-consumer lock-in effect}, Working paper, Cambridge Service Alliance (2014).
\newline\urlprefix\url{https://cambridgeservicealliance.eng.cam.ac.uk/system/files/documents/2014AugustPaperBusinesstoConsumerLockinEffect.pdf}

\end{thebibliography}

\clearpage
\pagenumbering{roman} 

\appendix

\section{Engineering \llmfkg{}}

\subsection{Ancillary Functions}
\label{ssec:ancillary}

\begin{algorithm}
    \captionsetup{font=\algCaptionSize}
    \caption{Adds an individual to a class, unless it is already assigned to a more specific class}
    \label{alg:add-individual}
    \begin{algorithmic}[1]\algCodeSize
        \Require $\relset{O} = \relset{C} \cup \relset{P} \cup \relset{X}$: partially populated ontology
        \Require $\instance{i}$ is an individual
        \Require $\concept{R} \in \relset{C}$: root concept $\instance{i}$ should be assigned to
        \Ensure $\relset{X}'$ contains the (possibly indirect) assignment of $\instance{i}$ to $\concept{R}$
        \medskip
        \Function{\addToClass}{$\relset{O}, \instance{i}, \concept{R}$}
            \If{$\exists \concept{C} \suchThat (\instance{i} : \concept{C}) \in \relset{X}' \wedge \concept{C} \sqsupset \concept{R}$}
                \State \Return $\relset{X} \cup \{ \instance{i} : \concept{R} \} - \{ \instance{i} : \concept{C} \}$
            \ElsIf{$\not\exists \concept{C} \suchThat (\instance{i} : \concept{C}) \in \relset{X}'$}
                \State \Return $\relset{X} \cup \{ \instance{i} : \concept{R} \}$
            \EndIf
            \State \Return $\relset{X}$
        \EndFunction
    \end{algorithmic}
\end{algorithm}

\llmfkg{} relies on a number of shared ancillary functions,
whose semi-formal definitions are provided here.

Functions \getRange{} and \getDomain{} are used to retrieve the domain and range of a given property,
respectively.
So, for instance, for property $\relation{p} \subseteq \concept{D} \times \concept{R}$, \Call{\getDomain}{$\relation{p}$}
returns $\concept{C}$, whereas \Call{\getRange}{$\relation{p}$} returns $\concept{R}$.

Function \askOracle{} is used to query LLM oracles,
whereas functions \extractBinary{} and \extractNames{} aim to extract binary answers,
or relevant individuals' or concepts' names from their textual responses,
respectively.
Their internal functioning is non-trivial,
and we discuss it in \ref{par:extract-names}.
Here,
we just focus on their syntax and expected semantics.
Accordingly, function \Call{\askOracle}{$\relset{L}, q$} accepts as input the LLM oracle $\relset{L}$ to query and the
query string $q$ to be submitted to the oracle, and it returns a string as response $text$.
Both the input ($q$) and output ($text$) strings are assumed to contain natural language text, and to be of arbitrary
length.
Function \Call{\extractNames}{$text$} accepts as input the response string $text$ from an LLM oracle, and it returns a
set of strings $names$, to be interpreted as names of individuals or concepts, depending on the concept.
Function \Call{\extractBinary}{$text$} performs a similar operation, but it returns a boolean value
(true/false, yes/no, \emph{et similia}).

Finally, function \addToClass{} is used to add individuals to classes.
Despite the simplicity of its intuitive definition, its actual operation is non-trivial, hence why we detail pseudo-code
in \Cref{alg:add-individual}.
In particular, we write \Call{\addToClass}{$\relset{O}, \instance{i}, \concept{C}$} to denote the assignment of
individual $\instance{i}$ to class $\concept{C}$ in ontology $\relset{O}$.
However, there may be corner cases in doing so, such as the case where $\instance{i}$ is already assigned to some
sub-class $\concept{S} \sqsubset \concept{C}$.
In that case, no actual edit is performed on the ontology.
Vice versa, if $\instance{i}$ is already assigned to some super-class $\concept{R} \sqsupset \concept{C}$, then
$\instance{i}$ is assigned to $\concept{C}$, and it is unassigned from $\concept{R}$.
In any case, the function returns a novel set of assertions $\relset{X}'$, which is possibly different from the input
one (denoted by $\relset{X}$).

\subsection{Practical Aspects}
\label{ssec:practical-aspects}

At the practical level, a number of aspects should be considered in order for \llmfkg{} to be effective.
Most notably, the abstract formulation from \cref{sec:contribution} models interaction with LLM oracles by means of
functions \askOracle{} and \extractNames{}, abstracting technicalities away.
However, in practice, the implementation of these functionalities is non-trivial, and it requires taking into account
technicalities involving
\begin{inlinelist}
    \item the engineering of prompts to be submitted to the LLM oracle,
    \item the mining of relevant individual or concepts' names from the textual responses of the LLM oracle,
    as well as
    \item corner cases involving empty or inconclusive responses,
    and
    \item other relevant aspects such as costs and rate limitations.
\end{inlinelist}

Accordingly, we here discuss each of these aspects in detail, following a means-oriented approach.

\subsubsection{Prompt and response engineering}

\begin{lstlisting}[
    float,
    basicstyle=\tiny\ttfamily,
    label={lst:cats},
    caption={Example of response from ChatGPT to the query \str{can you generate examples of cats?}}]
Certainly! While I can't physically generate images, I can certainly describe or provide information about different types of cats. Here are a few examples:

    1. Domestic Shorthair:
        Description: A common and popular cat breed known for its short, sleek coat.
        Characteristics: Versatile in colors and patterns, friendly, and adaptable.

    2. Siamese Cat:
        Description: Elegant and slender cat with distinctive color points.
        Characteristics: Vocal, social, and known for their striking blue almond-shaped eyes.

    (*@$\ldots$@*)

    10. Russian Blue:
        Description: Short-haired cat with a bluish-gray coat and striking green eyes.
        Characteristics: Reserved but affectionate, known for its plush double coat.

These are just a few examples, and there are many more cat breeds, each with its own unique characteristics and appearance.
\end{lstlisting}

Generally speaking, \llmfkg{} performs automatic queries to LLM oracles, with the purpose of inducing them to generate
individuals or concept names.
However, LLM come with no guarantee on the structure of the generated text, nor on its relevance w.r.t. the ontology:
the generated text may be long, unstructured, and it may contain both relevant and irrelevant chunks of information.
For instance, submitting the query \str{can you generate examples of cats?} to ChatGPT may produce a reponse similar to
the one in \cref{lst:cats}.
This may complicate the automatic extraction of relevant information from the textual responses of the LLM oracle,
as irrelevant information should be filtered out, while only relevant information should be retained.
In the example from \cref{lst:cats} the relevant information is the list cats' races' names, whereas the irrelevant
information is anything else.

There exists a delicate trade-off among the complexity of the prompt and the complexity of the response.
As a consequence, instead of engineering complex text extraction procedures aimed at distilling relevant information
from arbitrary text, one may try to engineer the prompt in order to induce the LLM oracle to generate more concise or
more structured responses.
In other words, a trade-off exists between the length / intricacy of query templates and the algorithmic complexity of
function \extractNames{}.

In the following paragraphs, we discuss a number of strategies to engineer the prompt in order to induce the LLM oracle
to generate more concise or more structured responses, and then we delve into the details of the \extractNames{}.
%

\begin{lstlisting}[
    float,
    basicstyle=\tiny\ttfamily,
    label={lst:cat-names},
    caption={Example of response from ChatGPT to the query \str{list of cat races, names only}}]
Certainly, here's a list of cat breeds with names only:

    1. Persian
    2. Siamese
    (*@$\ldots$@*)
    10. Domestic Shorthair
\end{lstlisting}

\paragraph{Getting concise and structured responses from oracles}
\label{par:concise-responses}


One may induce an LLM oracle to generate concise and structured responses by explicitly asking for these features
in the query.
For instance, at the time of writing, it is sufficient to ask for a \str{concise list} of something to let the LLM
generate text formatted as a list in Markdown format\footnotemark.
\footnotetext{\url{https://www.markdownguide.org/basic-syntax/\#lists-1}}

Moreover, we observe that complementing the query with the \str{names only} requirement, induces the LLM to
generate a very compact list of names, with no additional information.

Finally, we also observe that the word \str{concise} may be omitted when combining the two requirements.
We speculate that this is possible because the \str{names only} requirement already implies the maximum level of
conciseness.

As an example, consider again the case of \cref{lst:cat-names}.
There, the response was attained by submitting the query: \str{list of cat races, names only}.

Along this line, we recommend engineering query templates in such a way to include the aforementioned requirements.
So, for instance, individual seeking templates should be of the form \str{list of examples for \var{class}, names only},
whereas relation seeking templates should be of the form \str{list of examples of \var{property} for \var{individual},
names only}.
Similarly, best match templates should be of the form \str{best class for \var{individual} among \var{classes}, be
concise}.

\paragraph{Avoiding out-of-context responses}
\label{par:fancyname}


One may induce an LLM to generate context-aware responses by explicitly providing context in the query.
But what is context, precisely?

In the particular case of \llmfkg{} applied to a domain-specific ontology, the context is always (at least) two-folded.
In fact, the LLM oracle should be aware that queries may contain references to
\begin{inlinelist}
    \item the ontology's domain,
    and
    \item ontologies in general (e.g.\ ontologic jargon such as ``class'' or ``individual'').
\end{inlinelist}

In our running example about animals, the domain could be for instance ``a zoologist willing to populate an ontology
describing sorts of animals''.
This could be made explicit in the query by prepending a statement such as \str{you're a zoologist} or \str{zoological
context}.

The takeaway here is that query templates should contain explicit references to the ontology's domain, and possibly
explicitly clarify when they are seeking for individuals or class names.

Going back to our running example, consider for instance case of the query \str{can you name a few examples of cats?}.
This is very ambiguous, as no context is provided.
In fact, at the time of writing, ChatGPT replies with a list of common names for domestic cats, including;
\str{Whisker}, \str{Luna}, \str{Oliver}, \str{Cleo}, \str{Simba}, etc.
However, simply prepending the string \str{zoological context} to the query may induce ChatGPT to reply with a list of
members of Felidae family (e.g.\ \str{African Lion (Panthera leo)}, \str{Tiger (Panthera tigris)}, etc.) plus a
warning that \str{the term `cats' generally refers to members of the family `Felidae'}.
An even more precise result can be attained via the query \str{zoological context. can you name a few instances of class
`cat race'?}, where the ontological jargon is explicitly adopted.

In the specific case of \llmfkg{}, context information may be simply provided by means of ad-hoc text in query
templates.
Contextual text may be the same for all templates, or it may be slightly different for each template.
It could be simply prepended or appended to each template (e.g.\ \str{zoological context. list of examples for list of
examples for \var{class}, names only}).

Another key aspect to consider is that query templates are meant to be filled with class or property names from the
initial ontology definition.
This may be problematic, if the name of the class or property is ambiguous outside the context of the ontology.
To avoid this inconvenience, we assume that each class and property in the ontology may come with an optional
attribute, namely \relation{fancyName}, which is meant to be used in queries instead of the actual name of the
class/property.

Consider for instance the case of the aforementioned $\concept{Cat}$ class which may be present in the ontology, yet
ambiguous in the zoological domain.
Let us also assume that, for some technical / administrative reason, the class name cannot be changed.
When this is the case, the ontology designers may add the attribute \relation{fancyName} to $\concept{Cat}$, and set its
value to some zoologically-correct name, such as \str{feline species} or \str{felidae family}.
In this way, queries would be generated in an unambiguous way, despite the ontology contains a compact class name.

\paragraph{Mining relevant information from textual responses}
\label{par:extract-names}

\begin{lstlisting}[
    float,
    basicstyle=\tiny\ttfamily,
    label={lst:fake-cat-names},
    morekeywords={Persian,Maine,Coon,Caracal,Domestic,Shorthair},
    morendkeywords={Siamese,Bengal,Sphynx},
    caption={
        Example of LLM response containing a numbered list of fake cat names.
        Some items in the list actually contain more than one name.
        Some names are composed by more than one word.
        The response also contains text which is not part of the list.
        Relevant names are highlighted in blue and cyan (alternating colours).
    }]
Certainly, here's a list of cat breeds with names only:

    1. Persian or Siamese
    2. Maine Coon
    3. Bengal, Caracal
    4. Sphynx
    (*@$\ldots$@*)
    9. Domestic Shorthair

These are just a few examples, and there are many more cat breeds, each with its own unique characteristics and appearance.
\end{lstlisting}

Assuming that query templates are engineered in such a way to force LLM responses to contain a list of names
(spanning either a single line or multiple ones)
\llmfkg{} relies on function \extractNames{} for the localisation of those names in the response text.
To do so, the function attempts to parse LLM responses against a fairly flexible context-free grammar, tailored on the
localisation of relevant chunks of text.

Before delving into the details of the grammar and the parsing procedure, let us exemplify the typical situation.
Both in the population and relation phases, the LLM oracle is asked to produce a list of names, to be interpreted as
individuals.
Conversely, in the redistribution phase, the LLM oracle is asked to produce a single name, to be interpreted as a class,
among a set of candidate names which are selected by the algorithm.
So, in the general case, LLM is asked to produce a multitude of names, in the form of a list.
Empirically, we observed the LLM oracles tend to produce either multi-line lists or single-line lists.
Multi-line lists are typically formatted as bulleted or numbered lists in Markdown format, whereas single-line lists
are typically sentences containing commas, semicolons, or words such as \str{and} or \str{or}.
It is worth highlighting that responses containing singleton names, can be considered a trivial case of single-line
lists.
Finally, we observed that LLM oracles may sometime include irrelevant (w.r.t. \llmfkg{}) information in the response,
such as pleasantries, contextualisation text, or warnings about the fact that the response may be incomplete.
In \cref{lst:fake-cat-names}, we report a handcrafted response aimed at exemplifying one complex case, where inline and
multi-line lists are mixed, as well as irrelevant text.

The actual implementation of the \extractNames{} function is based on the assumption that meaningful response consists
of text (containing some sub-string) matching the following context-free grammar:
\begin{equation}\small
    \label{response-grammar}
    \begin{array}{rcl}
        \var{Response} & := & \var{SingleLine} \mid \var{MultiLine} \\
        \var{MultiLine} & := & \var{Item} (\var{Whitespace} \var{Item})* \\
        \var{Item} & := & \var{Bullet} \var{SingleLine} \\
        \var{SingleLine} & := & \var{Relevant} (\var{Separator} \var{Relevant})* \\
        \var{Relevant} & := & \var{Word} (\var{Whitespace} \var{Word})* \\
        \var{Word} &:= & \text{any combination of letters,} \\
        & & \text{punctuation, or digits} \\
        & \mid & \texttt{"} \var{Word} \texttt{"} \mid \texttt{`} \var{Word} \texttt{'} \\
        \var{Whitespace} & := & \text{any combination of spaces,}\\
        & & \text{tabulations, or newlines} \\
        \var{Separator} & := & \texttt{and} \mid \texttt{or} \mid \texttt{,} \mid \texttt{;} \\
        \var{Bullet} & := & \texttt{-} \mid \texttt{*} \mid \texttt{+} \mid \texttt{\#}+ \mid \var{Index} \texttt{.} \\
        \var{Index} & := & \texttt{1} \mid \texttt{2} \mid \texttt{3} \mid \dots \\
    \end{array}
\end{equation}
In the grammar, relevant snippets of text are denoted by the non-terminal symbol \var{Relevant}, which may consist of
one or more words, possibly separated by white space.
Different relevant snippets may be separated by a \var{Separator} -- which may be one of the following strings: \str{and}
or \str{or}, or \str{,} --, or they may be spanning multiple items of a bulleted or numbered list.

\begin{figure*}
    \centering
    \includegraphics[width=.8\linewidth]{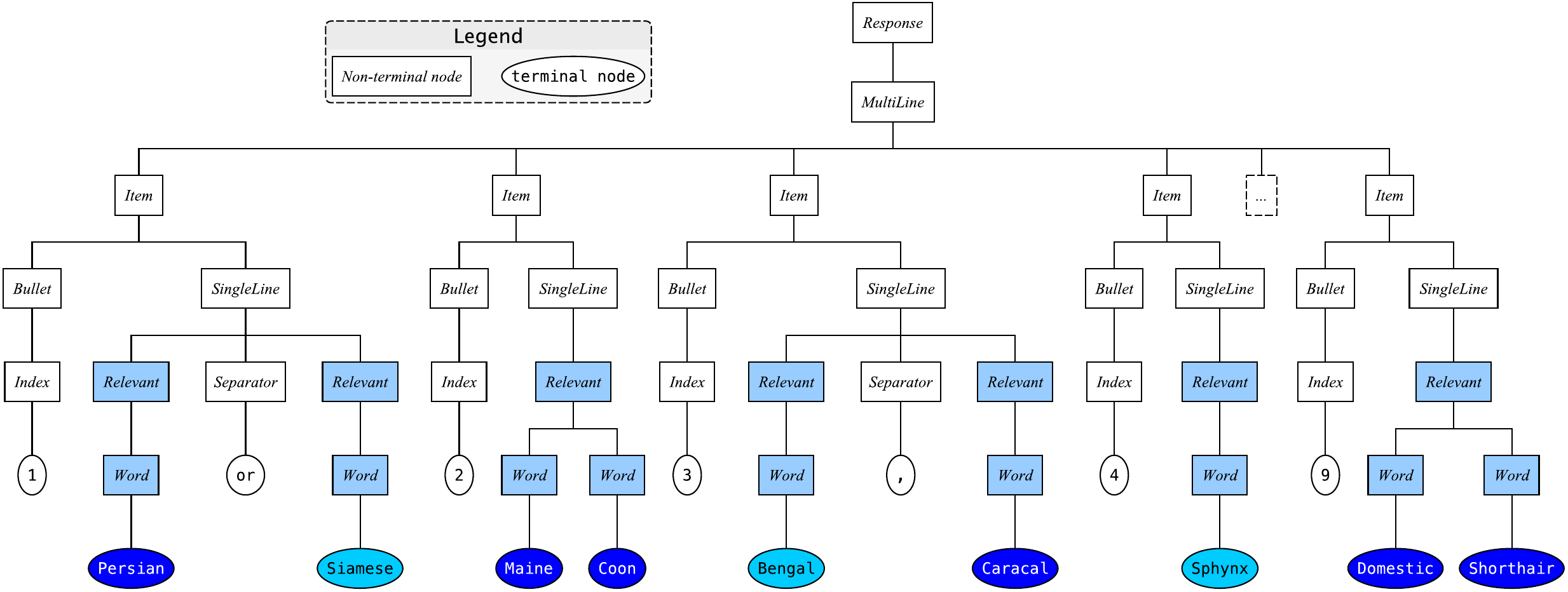}
    \caption{
        Parse tree for the response in \cref{lst:fake-cat-names}, parsed according to the grammar in \cref{response-grammar}.
        Colours highlight relevant sub-trees of the parse tree.
        Blue and cyan (alternating colours) ellipses highlights relevant names output by the \extractNames{} function
        when fed with that response.
    }
    \label{fig:parse-tree}
\end{figure*}

To extract relevant names from any given text, the \extractNames{} function would look for the longest sub-string
matching the grammar from \cref{response-grammar}, and, if any such sub-string exists, it would build a parse tree
for it.
For instance, for the response in \cref{lst:fake-cat-names}, the parse tree would look like the one in \cref{fig:parse-tree}.
In the parse tree, any node corresponds to non-terminal symbols in the grammar (rectangles in the figure), whereas
leaves correspond to terminal symbols (ellipses in the figure).
To select relevant names from the parse tree, the function would simply select all the subtrees rooted in a
\var{Relevant} node (coloured subtrees in the figure), and it would concatenate the leaves of each subtree to form
one relevant name to be output.
Notably, this procedure would automatically exclude irrelevant text from the response, there including any prologue
or epilogue which does not contain a list, as well as any irrelevant text in the middle of the list.

Technically, other fine-grained constraints may be imposed on the grammar, in order to make the extraction procedure
more robust.
For instance, one may require that the indexes of a numbered list are continuous and monotonically increasing, or that
bullets in bullet lists are all the same.
In the particular case of queries for the redistribution phase, one may also filter out from the response any relevant
name which is not included in the query (as that would imply the LLM is providing an answer which is not among the
admissible ones).

As far as the \extractBinary{} function is concerned, information mining is far less articulated.
In that case, the function simply looks for positive (`yes', `true', etc.) or negative (`no', `false') words
in the textual answer, in a case-insensitive way.
Combined with query templates which explicitly invite the LLM to be sharp in its answer,
this approach provides a simple yet effective way to interpret LLM responses as boolean values.

\paragraph{Handling empty or inconclusive responses}


Put it simply, the \extractNames{} function outputs a set of sentences, each one to be interpreted as a relevant name.
Despite all the precautions adopted in the design of the templates and in the parsing of responses, the function may
still output names that require post-processing.
For instance, it may happen that the LLM oracle politely refuses to provide an answer to the query.
When this is the case, the answer is not empty, but rather it contains a sentence such as \str{I'm sorry, but I'm a
large language model trained by OpenAI, and I \var{motivation\ here}}.
However, our parsing procedure would interpret this sentence as an inline list of names, such as \str{I'm sorry},
\str{but I'm a large language model trained by OpenAI}, etc., which are clearly meaningless.
This is the reason why any robust implementation of the \extractNames{} function should include a post-processing phase,
where irrelevant names are filtered out.
This can be done by means of a set of regular expressions, which are meant to match irrelevant snippets of text
in a sentence.
For instance, sentences containing sub-strings \str{I'm sorry}, \str{I'm a large language model}, etc. may be filtered
by the list of relevant names output by the \extractNames{} function.

As a consequence of this post-processing phase, the \extractNames{} function may output an empty list of names.
To mitigate this issue, users of \llmfkg{} may consider asking the same question multiple times in slightly different
way.
This is the reason why each phase of \llmfkg{} relies on \emph{a set} of query templates, rather than a single one.
Providing more than one query template for each phase is the way by which users can let \llmfkg{} ask the same question
multiple times, in slightly different ways.
To serve this purpose, it is quintessential that users of \llmfkg{} design query templates accordingly.

\paragraph{Regulating the creativity of the LLM oracle}

Despite the many tricks described so far, it is worth recalling that LLM oracles may still allow for a degree of
randomness when generating their responses.
Randomness is, to some extent, a desirable feature in this context as it allows for the generation of novel names,
yet it may also imply higher chances of generating meaning-less names.
Most commonly, LLM providers provide a way to regulate the randomness of their responses, by means of a parameter
called \emph{temperature} (cf. \cref{par:temperature}).

As far as \llmfkg{} is concerned, the temperature parameter can be set by the user when starting the algorithm, and that
value will be used for all queries submitted to the LLM oracle during a run of the algorithm.
It is worth mentioning that the algorithm will set temperature to 0.0
when querying the LLM oracle in the merging phase,
regardless of what the temperature the user has set.

\paragraph{About the syntactic similarity of individuals}

The merging phase heavily relies on the \synSimilar{} function,
which is in charge of determining whether two individuals names are similar enough to be considered for merging.
The implementation of this function is critical for the success of the merging phase,
as it is in charge of selecting candidates for merging.
Making it too permissive may lead the algorithm to do too many LLM queries,
whereas making it too strict may lead the algorithm to miss some duplicates.

We rely on a simple implementation where
\synSimilar$(\instance{i}, \instance{i})$ returns true if (and only if) the names of the two instances
$\instance{i}$ and $\instance{j}$ share an identical substring longer than $\lambda$ characters.
There, $\lambda$ is a hyper-parameter of \llmfkg{}, that we empirically set to $4$.

\subsubsection{Costs and rate limitations}
\label{ssec:costs-rates}

As LLM text generation functionalities are currently provided `as a service', providers may charge users for the
queries they submit to the LLM oracle, or they may limit the rate at which queries can be submitted.
On the \llmfkg{} side, this may be problematic, as the number of queries submitted to the LLM oracle may be very high.
Hence why any practical implementation of \llmfkg{} should be robust w.r.t. abrupt service unavailability, and it
should minimise the financial cost of the queries submitted to the LLM oracle.
In the following paragraphs, we discuss a number of strategies to achieve these goals.

\paragraph{Automatic query rate regulation}


Rate limitations are typically perceived one the user side as service disruptions: the LLM service provider may simply
reject queries because too many queries have been submitted in a short time span---or maybe because too many
tokens have been consumed/produced by the previous queries and the corresponding answers.

To mitigate this issue, practical implementations of \llmfkg{} should be able to automatically regulate the rate at
which queries are submitted to the LLM oracle.
One simple yet effective strategy to achieve this goal is the so-called \emph{exponential back-off} strategy---which is
commonly used in communication protocols to avoid congestion.

The exponential back-off strategy is based on the assumption that a failure to submit a query to the LLM oracle may most
likely due to a rate-limitation or service-overload situation, and that retrying the query after a short delay may be
sufficient to overcome the issue.
If it is not, the same strategy could be applied again, with a longer delay.


For instance, as LLM provides' rate limitations are typically expressed in terms of queries or tokens per minute,
one may consider setting the initial delay to $30\text{s}$ and the back-off factor to $150\%$.
In this way, the first time a query is submitted, there is no delay, the second attempt would be delayed of 30 seconds,
the third one of 45 seconds, the fourth one of roughly more than one minute, and so on.
Of course, one may also consider limiting the amount of retries, or to set a maximum cumulative delay.

In \llmfkg{}, the exponential back-off strategy is implemented by means of the \askOracle{} function, which is
responsible for submitting queries to the LLM oracle.
%

\paragraph{Minimising costs}

Minimising financial costs can either be achieved by minimising the number of queries submitted to the LLM oracle, or by
minimising the amount of tokens involved in each query.
Notice that the second option may imply the first option: fewer tokens produced per query implies fewer names being
generated by \llmfkg{}, which implies fewer individuals are added to the ontology, which implies fewer queries are
submitted in the relation and redistribution phases.
Accordingly, we here discuss how to minimise the amount of tokens consumed and produced by each query.

Our first hint is to design query templates in such a way to suggest the LLM to generate short and structured responses,
as already discussed in \cref{par:concise-responses}.
However, that would only reduce the \emph{produced} tokens.
To minimise the \emph{consumed} tokens, one may consider making query templates, class names, property names, and fancy
names as concise as possible.
So, for instance, the query template \str{list of examples for \var{class}, names only} is sub-optimal, as it contains
as it could be shrank to \str{examples list for \var{class} names only}, with no relevant change in the generated
response.

A more direct way to impose token limitations is to exploit ad-hoc API options provided by LLM providers, if any.
For example, the OpenAI allows clients to specify the maximum amount of tokens in the response via the
query parameter `\texttt{max\_tokens}'\footnotemark.
\footnotetext{\url{https://platform.openai.com/docs/api-reference/chat/create\#max\_tokens}}.
When such parameter is set, the LLM oracle will stop generating text as soon as the maximum amount of tokens is reached.
This may result in truncated responses, but it may give full control on the amount of tokens consumed by each query%
---therefore making financial planning easier.

It is worth remarking that truncated responses may require further post-processing, especially as far as names
extraction is concerned.
When truncation is in place, there is no guarantee that the last extracted name is complete, as therefore it should be
dropped---in order to avoid introducing corrupted data into the ontology.

Another feature that may be exploited to minimise costs in face of trials and errors -- which are almost unavoidable
when fine-tuning query templates on a new ontology -- is \emph{caching}.
By caching queries (along with their parameters such as temperature and max tokens) and the corresponding answers, one
may avoid costs in case the same query is submitted again.
This is not really useful within the single run of \llmfkg{}, but it may be useful to avoid costs when re-running
\llmfkg{} on the same ontology, or when re-running \llmfkg{} on a similar ontology.
Moreover, it is possible to exploit caches for debugging purposes, e.g.\ for retrospectively analysing an unsatisfying
ontology population attempt.
This could also be useful to compare different runs of \llmfkg{} on the same ontology, or on different ontologies.

\end{document}